\documentclass[journal]{IEEEtran}
\usepackage{cite}
\usepackage{graphicx}
\usepackage{subfig}
\usepackage{url}
\usepackage{comment}
\usepackage{todonotes}
\usepackage{amsmath}
\usepackage[utf8]{inputenc} 

\begin{document}

\title{Automated Playtesting with Procedural Personas through MCTS with Evolved Heuristics}
\author{Christoffer Holmg\aa rd$^1$~\IEEEmembership{member,~IEEE}, Michael Cerny Green$^2$, Antonios Liapis$^3$~\IEEEmembership{member,~IEEE}, Julian Togelius$^2$~\IEEEmembership{member,~IEEE}\\
1: Department of Art + Design, Northeastern University\\
2: Tandon School of Engineering, New York University\\
3: Institute of Digital Games, University of Malta
}

\maketitle

\begin{abstract}
This paper describes a method for generative player modeling and its application to the automatic testing of game content using archetypal player models called procedural personas.
Theoretically grounded in psychological decision theory, procedural personas are implemented using a variation of Monte Carlo Tree Search (MCTS) where the node selection criteria are developed using evolutionary computation, replacing the standard UCB1 criterion of MCTS.
Using these personas we demonstrate how generative player models can be applied to a varied corpus of game levels and demonstrate how different play styles can be enacted in each level.
In short, we use artificially intelligent personas to construct synthetic playtesters.
The proposed approach could be used as a tool for automatic play testing when human  feedback is not readily available or when quick visualization of potential interactions is necessary. 
Possible applications include interactive tools during game development or procedural content generation systems where many evaluations must be conducted within a short time span.
\end{abstract}


\section{Introduction}\label{sec:intro}
One of the challenges faced by game designers is predicting how different players will interact with the the systems and content that they are crafting. Most games are complex emergent systems that allow for a variety of interaction patterns, depending on the player's preference(s) and the interaction between player, game, and any other players.
Game designers employ a variety of processes to imagine and observe how different types of players might respond to their content. The processes can be thought of as existing on a spectrum, ranging from the designer imagining what different types of players might do, to analyzing play data of beta testers or a portion of the player base in the case of continuously updated ``live'' games. Each approach has different strengths, weaknesses and costs, making it relevant for different game makers or different stages of the game making process \cite{fullerton2004game,el2013game}.

\begin{figure}[hbt]
\centering
\includegraphics[width=0.95\columnwidth]{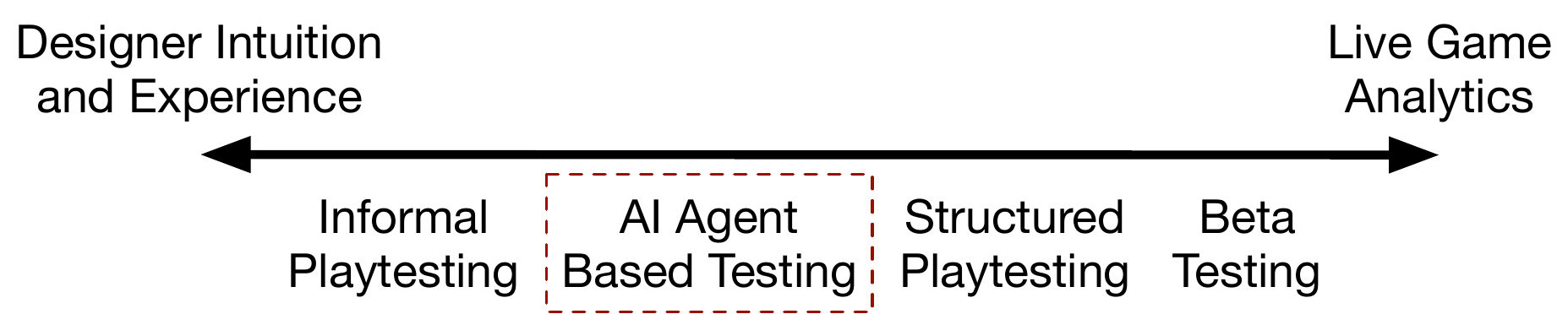}
\caption{Spectrum from simple to complex design testing methods in game development.} 
\label{fig:testing_spectrum}
\end{figure}

In this paper we suggest and demonstrate a method taking a new position on this spectrum (illustrated in Figure~\ref{fig:testing_spectrum}): using archetypal generative player models as critics for game content, enabling automated playtesting and evaluation of game content; here specifically levels. We identify this approach as the use of \emph{Procedural Personas for Playtesting}.

We evolve artificially intelligent game playing agents for the game \emph{MiniDungeons 2} \cite{holmgard2015minidungeons2}.
The agents, or \emph{personas}, are characterized by different utility functions for their decision-making.
These utility functions capture various archetypal goals that players might hold in relation to the affordances and potential interactions of the particular game.
To control the personas, we use a variant of Monte Carlo Tree Search (MCTS) which is well-suited to building biased search trees in large search spaces.
However, rather than applying the Upper Confidence Bound 1 applied to Trees (UCB1) formula typically used for MCTS, we use genetic programming to evolve persona-specific evaluation formulas.
This allows us to find mappings between persona utility functions and state evaluation algorithms.
We evolve well-performing game-playing agents for all defined personas through this variant of MCTS.

Using the evolved personas, we show how different levels can be automatically evaluated in terms of their playability for players holding different preferences.
This approach can be useful to game creators as it provides an insight into dynamic properties \cite{hunicke2004mda} of their content as they are crafting the mechanics.
For instance, such agents could eventually be added directly to a game engine's editor to allow for almost real-time feedback during content creation.
The approach can also be used as an automated evaluation mechanism for procedural content generation of game content, where procedural personas can function as stand-ins for the game designer or different players when large amounts of content have to be evaluated.

This paper builds on our previous work on MCTS agents with hand-crafted utility scores for the MiniDungeons 2 game \cite{holmgard2015mcts}, expanding on those concepts by broadening the number and utility of personas (through human design) as well as discovering UCB-like criteria (through evolution) which outperform UCB1. More broadly, this paper enhances our earlier definitions of procedural personas \cite{holmgard2014generativeagents,holmgard2014evolvingpersonas,holmgard2015evolving} which were applied for simulation-based level generation \cite{liapis2015personacritics}. However, the MCTS agents used in this paper are more modular in their utility definitions and afford far faster runtimes when performing automated playtesting. Moreover, the MCTS agents in this paper are tested on MiniDungeons 2, a far more complex game than its predecessor MiniDungeons introduced in \cite{holmgard2014generativeagents}.



\section{Related Work}\label{sec:related_work}
The approach taken in this paper draws on psychological decision theory, persona theory from design research, and player modeling and agent control from computational intelligence.
The procedural persona approach draws all of these four strands of work together in one framework for automatic playtesting in order to create generative player models that to some extent decide and play like human players.
This section briefly covers some of the foundational areas before describing prior work in bringing these approaches together.

\subsection{Personas for Game Design}\label{sec:play_personas}
The use of computational methods to imbue computer game characters with personality has been a focus of game AI programming since the very beginning of the medium.
As one instance, Short provides an overview of how non-player characters can be provided with human-like personalities under the heading of procedural personalties \cite{short2016procedural}.

The use of personas has a long history within design in general and design for information technology in particular.
The approach was pioneered for software development in the early 1990s \cite{cooper2004inmates} as a method for structuring and operationalizing qualitative data gathered from design research, chiefly in the form of interviews.
Based off interview data, a number of personas would be defined.
Each of these would serve as a specific instantiation of groups of user concerns that tended to co-occur, expressed as an archetypal example user, fully fleshed out with names, back stories, concerns and preferences.

Canossa and Drachen transported this approach into the realm of game design \cite{canossa2009patterns}, defining personas less in terms of general life concerns and more in terms of player interaction preferences within the space of the game. They call this new conceptualization \emph{play personas}
and operationalized their definition through data mining, suggesting how the persona design process could be supported by analyzing quantitative game data gathered via telemetrics \cite{tychsen2008defining}.
While play personas are archetypal models of player behavior inferred from experience or observed data, the re-projection into the game itself is something that is done imaginatively by the designer(s) of the game: i.e. play personas let us understand what players \emph{have done}, but do not enact what players \emph{might do}.
Procedural personas~\cite{holmgard2015evolving,holmgard2014generativeagents,holmgaard2014personas} extend the play persona idea by adding a game-playing, generative aspect.
By capturing persona characteristics from designer specification or from observed data, and formalizing these as utility functions, procedural personas are implemented as agents that can act in the game, enabling automatic playtesting.

Other work in the literature has investigated game testing without natural player data, notably \cite{nelson2011game}.
The approach taken here differs from the approach taken in \cite{nelson2011game} as it focuses on generating data from simulated players, rather than taking into account a larger number of potential metrics where some are not centered on player actions. As such, the procedural persona concept is a specialized case of \emph{player modeling}.

The line of work leading to this paper has been inspired by the category of ``Generative Action Models'' in the survey on player modeling by \cite{smith2011inclusive}. Until recently, this particular category has been underpopulated.

\subsection{Player Modeling}\label{sec:player_modeling}
Player modeling is the learning and use of computational models of player preference, experience and/or behavior~\cite{yannakakis2013playermodeling}.
Procedural personas, as generative player models, cover some of these aspects: behavior and preferences.
Other work in player modeling take different approaches to modeling play behavior and preference generatively. Perhaps the most obvious approach is to use some form of supervised learning to derive a model from play traces~\cite{ortega2013imitating,togelius2007towards}.
Cowley \emph{et al.} developed the concept of \emph{behavlets}, features of play derived from observed action sequences, structured through a top-down application of psychological temperament theory combined with machine learning \cite{cowley2016behavlets}. While behavlets can be used as generative models, they do not allow for the specification of player motivations without observations.
In contrast, procedural personas are driven by utility functions that can be either specified in a top-down manner by game or persona designers or formulated from play data through methods like inverse reinforcement learning \cite{tastan2011learning} or evolution \cite{holmgaard2014personas}.
The particular agent control method that is used to formulate a policy for procedural personas is technically arbitrary, as long as it can accept a utility function as a method of evaluation.
The most appropriate method may depend on the game for which the personas are being implemented. Prior work has shown that evolutionary methods and MCTS have potential for defining personas for turn based roguelike games \cite{holmgard2014evolvingpersonas,holmgaard2014personas,holmgard2015mcts}.

Devlin \emph{et al.} showed how observations of human play data can be used to bias MCTS to play the card game Spades \cite{devlin2016combining}. They use a relative entropy measure to assess the similarity of playing styles to traces of human players.
Zook \emph{et al.} limited the computational resources of MCTS to simulate player skill for a number of games \cite{zook2015monte} and similar findings were reported by Nelson \emph{et al.} \cite{nelson2016investigating}.
Another approach to biasing the MCTS search process to be more similar to human players is described by Khalifa \emph{et al.}~\cite{khalifa2016modifying}.
In the study described here, we take a similar approach and implement a variation of MCTS. We bias the search using evolution applying designer-defined utility outcomes as the fitness function.

\section{Monte Carlo Tree Search}
MCTS has shown considerable potential and flexibility as a game-playing algorithm \cite{browne2012mcts}.
For our purposes, MCTS has several desirable properties which approximate how decision making occurs in humans
: it evaluates the next best action based on a utility score for a predicted future state and operates under uncertainty of future outcomes. It also seems that by giving an MCTS algorithm more or less resources, you can simulate strategic depth in human decision-making~\cite{zook2015monte,nelson2016investigating}.

\subsection{Traditional MCTS}\label{sec:mcts}
As discussed in Section \ref{sec:intro}, MCTS is a tree search algorithm which creates biased search trees for decision processes. Unlike other tree search algorithms like Minimax, breadth-first, or depth-first, MCTS focuses on \emph{exploiting} the most promising moves to expand next, while balancing that by \emph{exploring} more neglected branches of the tree. The balance of exploitation versus exploration is traditionally handled through the evaluation of the Upper Confidence Bound for Trees equation, which applies UCB1 to the tree \cite{browne2012mcts}. The tree is built incrementally, with each iteration following a simple formula:

1. \textbf{Selection}: MCTS chooses the next node to expand via the \textit{tree policy}, starting at the root node and recursively picking the highest scoring child ``until the most urgent expandable node is reached'' \cite{browne2012mcts} or a terminal state (i.e. the game is won or lost). The score for traversing the tree in MCTS is termed \emph{tree policy}, and in traditional MCTS approaches it is given by the Upper Confidence Bound (UCB1) equation:
\begin{equation}
UCB = \frac{w_i}{n_i} + c{\cdot}\sqrt[]{\frac{\ln(t)}{n_i}}
\label{eq:ucb}
\end{equation}
where $w_i$ is the number of wins which originate from taking move $i$, $n_i$ is the times move $i$ was visited, $c$ is the exploration constant. It is typically chosen that $c=\sqrt[]{2}$, since this value has been shown to guarantee convergence to a value function within finite time for single-player games terminal states and rewards bounded to the range $[0,1]$\cite{browne2012mcts}. 
$t$ is the total number of simulations for the node considered and is equivalent to the sum of all $n_i$ for all possible moves. The UCB1 equation attempts to balance exploration (looking into paths not yet simulated) and exploitation (looking into paths previously simulated that show good results).

2. \textbf{Expansion}: unless the selected node is a terminal state (i.e. the game is over), a child node ($W$) is created for the next action. Typically, this next action is randomly selected from all possible future actions.

3. \textbf{Simulation}: the \emph{default policy} is used to simulate a random rollout from $W$. The rollout consists of performing actions at random until the game reaches a terminal state, or up to a fixed number of moves.

4. \textbf{Backpropagation}: the result (i.e. utility score) of the simulation is backpropagated to every node, from the expanded $W$ to the root node. This affects future policy decisions, i.e. future selection steps.

These four steps are applied sequentially until the computational resources allocated for the agent's move are depleted. The agent then chooses the next move (i.e. the child of the root node) with the highest utility score.

\subsection{Evolutionary Tree Policy}\label{sec:mcts_evolution}

As discussed in Section \ref{sec:mcts}, selection in MCTS must balance between exploitation and exploration; UCB1 is traditionally used to maintain this balance. Changes to the UCB1 formula of eq.~\ref{eq:ucb} are typically done in order to optimize it for a certain kind of game or to weigh certain kinds of game-play differently~\cite{khalifa2016modifying}.
Cazenave's work \cite{cazenave2007evolving} on evolving UCB1 alternatives for Go MCTS agents demonstrated that the resulting agent significantly outperformed peers that used traditional UCB equations, or even agents that used UCB1 alternatives specifically created for Go. In the General Video Game AI (GVGAI) framework, Bravi \emph{et al.} explored the possibility of evolving UCB1 replacements that were not specialized for one particular game \cite{bravievolving}. In this work, we use the approach of Bravi \emph{et al.} not to specialize the agent for particular games, but to bias its playstyle.


\section{The MiniDungeons 2 Game}

MiniDungeons 2 is a deterministic, turn-based rogue-like game, first described in \cite{holmgard2015minidungeons2}, in which the player takes on the role of a hero traversing a dungeon level, with the end goal of reaching the exit. The game stage is set on a 10 by 20 tile grid. Each tile is either an impassible \textit{wall} or a passable \textit{floor}. Floor tiles may contain objects that the hero can interact with, game characters such as the hero or non-player characters (NPCs), or nothing at all. Gameplay objects come in several varieties such as treasures, potions, portals, traps, and the exit of the level. To win, the player must reach the exit. All game characters have Hit Points (HP) and may deal damage. The player starts with 10 HP; the player loses when they run out of HP and die. 
Movement within the game is fairly simple. The player gets the first move every turn, and all NPCs move after.
The NPCs move in turn according to their original position on the map, starting from the top left corner and moving row-wise left-to-right until the bottom right corner is reached. This initial move sequence is retained even if NPCs later move to other locations.
On their turn, a game character may move in one of the four cardinal directions (North, South, East, West) so long as the tile in that direction is not a wall.
The player is given one re-usable javelin at the start of every level.
The player may choose to throw this javelin and do 1 damage to any monster within their unbroken line of sight.
After using the javelin, the hero must traverse to the tile to which it was thrown in order to pick it up and use it again.
A map contains many different objects the player can collide with. Different objects have different effects:
\begin{itemize}
\item\textbf{Potions} are used to increase the HP of the hero by 1, up to the maximum of 10. When collided with by either the hero or blobs, they are consumed and may not be re-used.

\item\textbf{Treasures} are used to increase the treasure score of the hero.  When collided with by either the hero or ogres, they are consumed and may not be re-used.

\item\textbf{Portals} come in pairs. If the hero collides with a portal, they are immediately (on the same turn) transported to the other paired portal.

\item\textbf{Traps} deal 1 damage to any game character moving through them, every time.
\end{itemize}
While exploring a map, the hero may come across a variety of monsters, some of which have  secondary goals in addition to attacking the player.

\begin{itemize}
\item\textbf{Goblins} (or Melee Goblins) move 1 tile every turn towards the player if they have an unbroken line of sight to the player. They have 1 HP and deal 1 damage upon collision. Goblins avoid colliding with other goblins and goblin wizards.

\item\textbf{Goblin Wizards} (or Ranged Goblins) cast a spell at the hero if they have an unbroken line of sight within 5 tiles of the player that does 1 damage. If they are over 5 tiles from the player but have line of sight, they will move 1 tile towards the player. Wizards have 1 HP and deal no damage on collision.

\item\textbf{Blobs} do not move unless they have unbroken line of sight with either a potion or the hero. If they see either one, they will move 1 tile towards the closest one per turn, preferring potions over the hero in case of a tie. A blob colliding with a potion consumes it. Blobs colliding with other blobs merge into a more powerful blob. The lowest level blob has 1 HP and does 1 damage upon collision. The 2\textsuperscript{nd} level blob has 2 HP and does 2 damage. The most powerful blob has 3 HP and does 3 damage. 

\item\textbf{Ogres} also do not move unless they have unbroken line of sight with either a treasure or the hero. If they see either one, they will move 1 tile towards the closest per turn, preferring treasures over the hero in case of a tie.  When an ogre collides with a treasure, they consume it, and their sprite becomes fancier to look at. Ogres have 2 HP and deal 2 damage to anything they collide with, including other ogres.

\item\textbf{Minitaurs} always move 1 step along the shortest path to the hero as determined by A* search, regardless of line of sight. Collision with the minitaur will deal 1 damage. A minitaur has no HP and is incapable of dying. If damage is done to it, the minitaur will be knocked out for 3 rounds (and can be passed through).
\end{itemize}
The game is technically infinite with all current maps as they all contain areas where the player might move back and forth, continuously dealing with the Minitaur using the Javelin.
However, in practice most maps are finished in 20-30 moves with goal directed play.
The branching factor is estimated to 3.41 across the included maps \cite{holmgard2015minidungeons2}, but depends on the map.

\section{Procedural Personas in MiniDungeons 2}\label{sec:method}

We identified four player archetypes which will become our procedural personas.
These personas prioritize different interactions with the game and were defined from the four primary objects in the game. The following four archetypes were defined based on our design experience and intuition: 
\begin{itemize}
\item \emph{Runner} aims to reach the exit.
\item \emph{Monster Killer} wants to kill monsters.
\item \emph{Treasure Collector} desires to collect treasure.
\item \emph{Completionist} attempts to consume any game object that can be collected or killed (monsters, potions, treasures).
\end{itemize}
Apart from the Completionist, these personas have been also featured in previous attempts at modeling personas via MCTS \cite{holmgard2015minidungeons2} or in the simpler {MiniDungeons} game \cite{holmgard2014evolvingpersonas,holmgard2014generativeagents}. Since the types of interactions with the game world are limited, these four single-minded personas capture a large part of the potential play space in MiniDungeons 2.

The personas all use MCTS to formulate a sequence of actions for play. Because MiniDungeons 2 is fully deterministic, each persona only builds one tree per map. It will immediately cease construction once a winning terminal state is discovered or it reaches timeout, wherein it will take the best sequence of actions it discovered. On average, trees will contain between two and five million nodes.
\subsection{Utility Formation}\label{sec:utility}

\begin{table} [t]
\caption{Gameplay metrics used as variables combined in the evolving equation trees, and their notation.}\label{table:equationSymbols}
\centering
\begin{tabular}{|l||l|}
\hline
Steps Taken (\texttt{ST}) &  Proximity to Exit  (\texttt{PE}) \\
Potions Drunk (\texttt{PD}) & Treasures Opened  (\texttt{TO})\\
Minitaur Knockouts  (\texttt{MTK}) & Monsters Slain (\texttt{MS})\\
Javelins Thrown  (\texttt{JT}) & Health Left  (\texttt{HL})\\
Teleports Used  (\texttt{TU}) & Traps Spring  (\texttt{TS}) \\
Average MCTS reward  (\texttt{\={R}}) & Interactive Objects Consumed (\texttt{IC}) \\
\hline
\end{tabular}
\end{table}

The four personas of MiniDungeons 2 are defined by their primary and secondary objectives, which calculate a move's utility score at the end of a simulation phase and is back-propagated to the rest of the tree in the next phase (see Section \ref{sec:mcts}).
From preliminary experiments, it seems that MiniDungeons 2 maps are frequently too complex and long for MCTS to simulate rollouts to a terminal state, as games can become infinite if the hero moves back and forth in place.
Therefore, in the rollout stage, our agents simulate 10 random moves before back-propagating the utility score. The different personas use metrics collected from the game's state after 10 random moves: Table 1 describes the variable metrics used in this paper. Note that for $PD$, $MS$, $TO$, and $IC$ the values represent the ratio out of all potions, monsters, treasures, and all non-monster game objects in the map, respectively.

\textbf{Runner (R)} has the primary objective of finding the exit in the fewest moves possible.  
\begin{equation}
U_R = 
\begin{cases}
PE - 0.01 \cdot ST  	&\text{if hero is alive}\\
PE - 0.01 \cdot ST - 5 	&\text{if hero is dead}\\
\end{cases}
\label{eq:baseline_r}
\end{equation}

\textbf{Monster Killer (MK)} has the primary objective of killing as many monsters as possible with the secondary objective of finding the exit.  
\begin{equation}
U_{MK} = 
\begin{cases}
0.7\cdot MS + 0.3\cdot PE &\text{if hero is alive}\\
0.7\cdot MS + 0.3\cdot PE - 5 &\text{if hero is dead}\\
\end{cases}
\label{eq:baseline_mk}
\end{equation}

\textbf{Treasure Collector (TC)} has the primary objective of consuming as much treasure as possible with the secondary objective of finding the exit.  
\begin{equation}
U_{TC} = 
\begin{cases}
0.7\cdot TO + 0.3\cdot PE &\text{if hero is alive}\\
0.7\cdot TO + 0.3\cdot PE - 5 &\text{if hero is dead}\\
\end{cases}
\label{eq:baseline_tc}
\end{equation}

\textbf{Completionist (C)} has the primary objective of consuming as many potions and treasures, and killing as many monsters as possible (thus ``completing'' a map), along with the  secondary objective of reaching the exit.
\begin{equation}
U_C = 
\begin{cases}
0.7\cdot IC + 0.3\cdot PE &\text{if hero is alive}\\
0.7\cdot IC + 0.3\cdot PE- 5 &\text{if hero is dead}\\
\end{cases}
\label{eq:baseline_c}
\end{equation}
By studying how these personas traverse the game's maps, we can better evaluate how players will interact with the game.

\subsection{Evolving the Policy of Personas}\label{sec:evolving_method}
Genetic programming is used to evolve the mathematical formula that replaces UCB1.
The Evolute C\# source code\footnote{\url{http://evolute-csharp.sourceforge.net}} was modified to work as follows:
The chromosome representation is a syntax tree where all nodes contain a \textit{binary} operation and all leaves contain a \textit{variable} or a \textit{constant}. The four binary functions are addition, subtraction, multiplication, and division. Constant values are uniformly, randomly generated floats within $[-1,1]$.  Variable values are derived from the game-play metrics described in Table \ref{table:equationSymbols}. The generator takes these variables and constant numbers and initializes equation trees with them, with an initial minimum depth of 2 and maximum of 5. During evolution, the maximum depth of a tree is set to 8 to avoid extremely long equations. This means that a equation can have as many as $2^8$ (i.e.~256) elements.

Each persona has its own utility function, as per Eqs.~(\ref{eq:baseline_r}--\ref{eq:baseline_c}), which evaluates the result to be back-propagated after the simulation step of MCTS. To test the candidate function, the UCB equation is completely replaced by the candidate as the tree policy. The agents are evaluated based on a fitness function calculated at the end of each playthrough, i.e. when the hero has reached the exit, when the hero is killed, or after a maximum allocated time has passed. Each persona uses the same fitness as the utility score (e.g. $f_{MK}=U_{MK}$) calculated at the end of the playthrough. E.g. for the Monster Killer the fitness $f_{MK}$ evaluates how many monsters it killed in total in this map, how close it ended to the exit ($PE=0$ if the exit was reached), and whether the playthrough ended because the hero was killed (which applies a penalty to the fitness).

\begin{figure*}[t]
\centering
\subfloat[Map 1]{\includegraphics[width=40px]{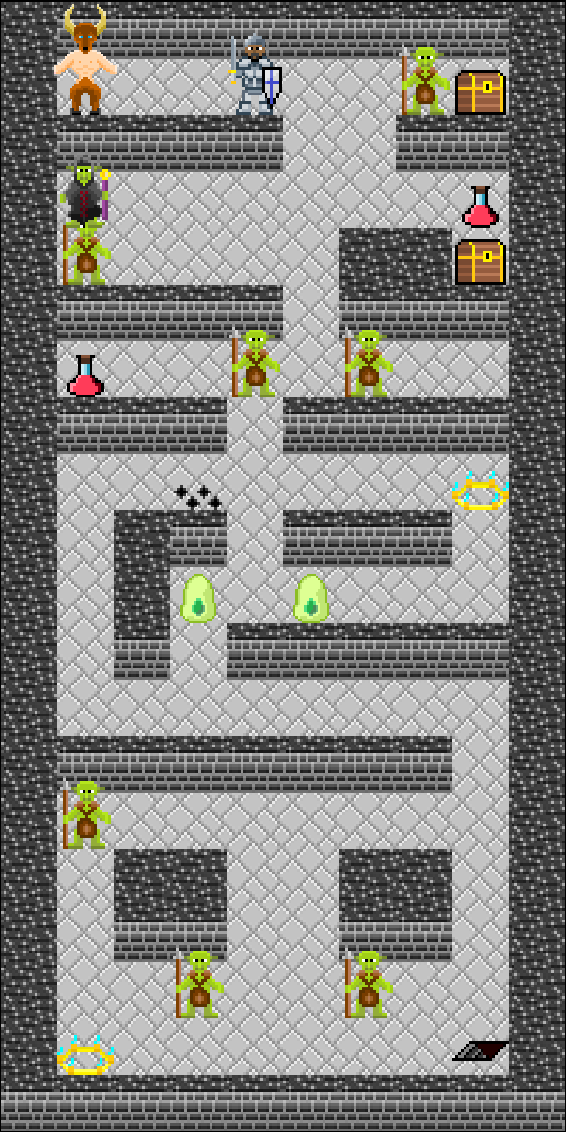}}~
\subfloat[Map 2]{\includegraphics[width=40px]{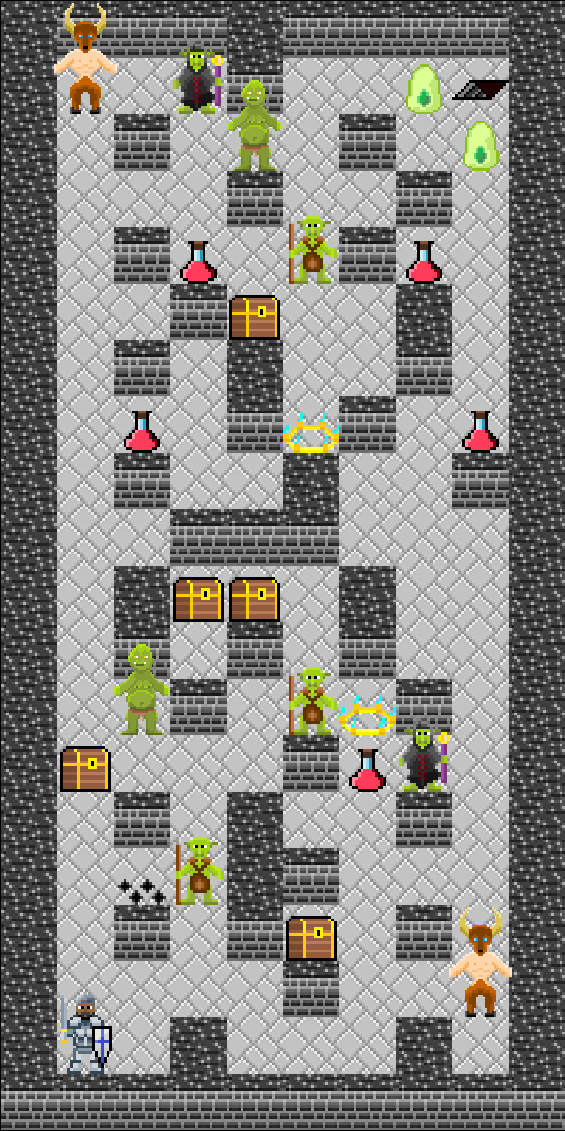}}~
\subfloat[Map 3]{\includegraphics[width=40px]{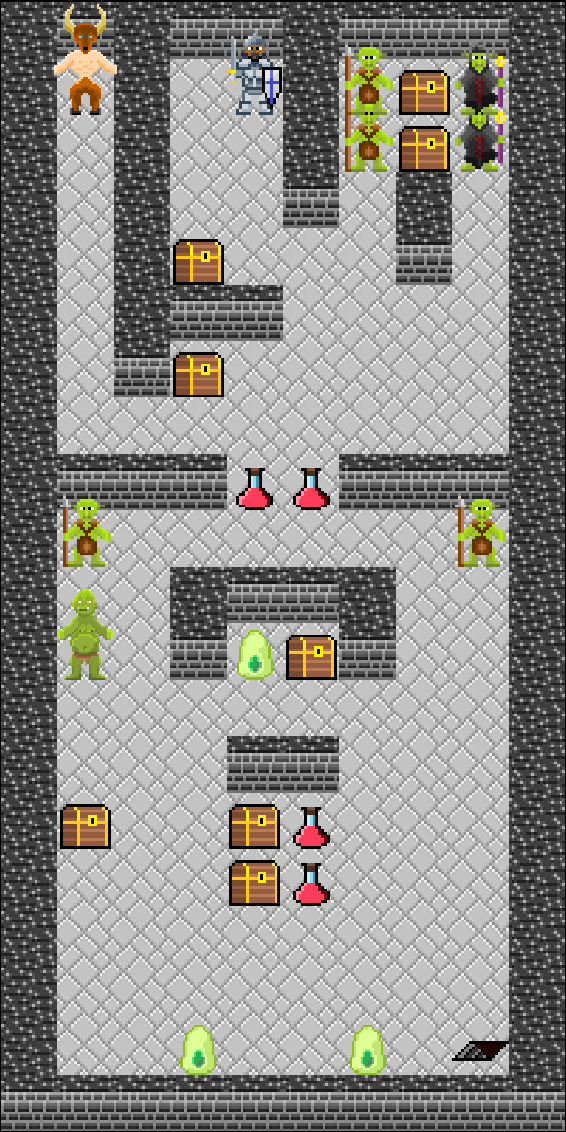}}~
\subfloat[Map 4]{\includegraphics[width=40px]{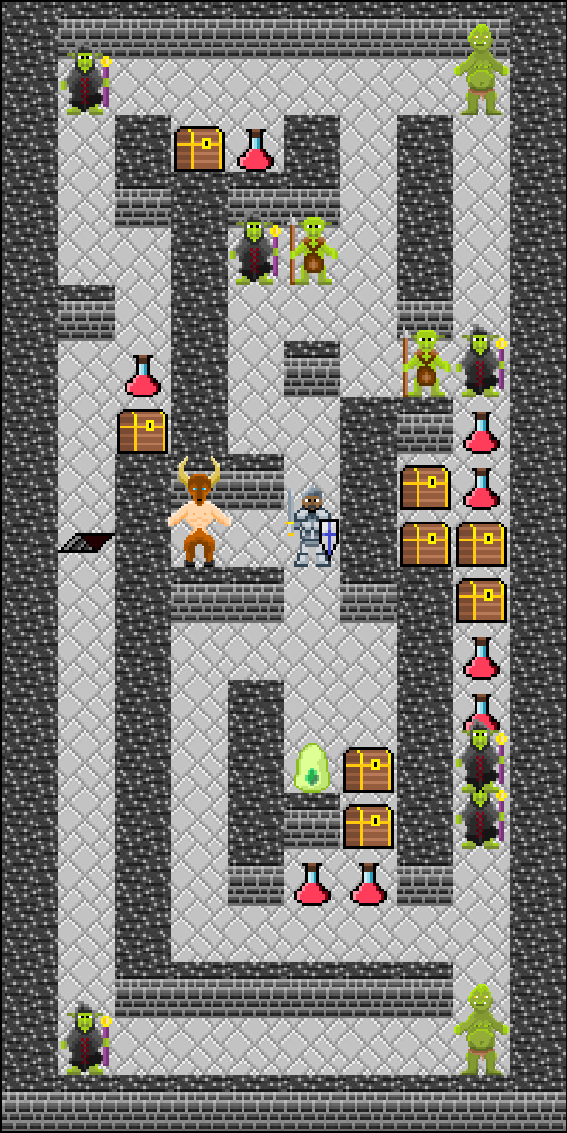}}~
\subfloat[Map 5]{\includegraphics[width=40px]{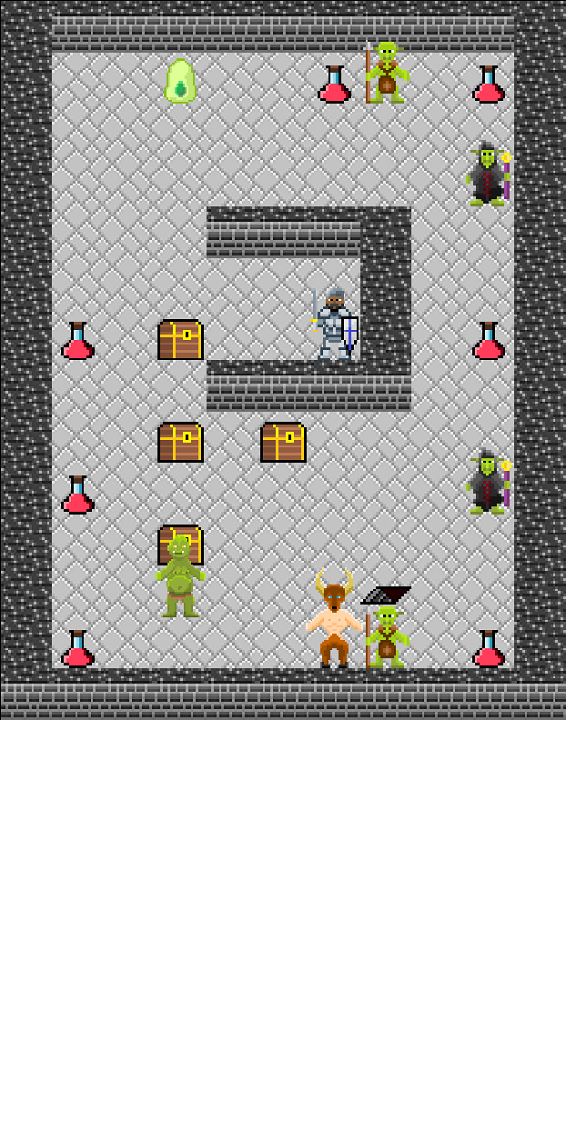}}~
\subfloat[Map 6]{\includegraphics[width=40px]{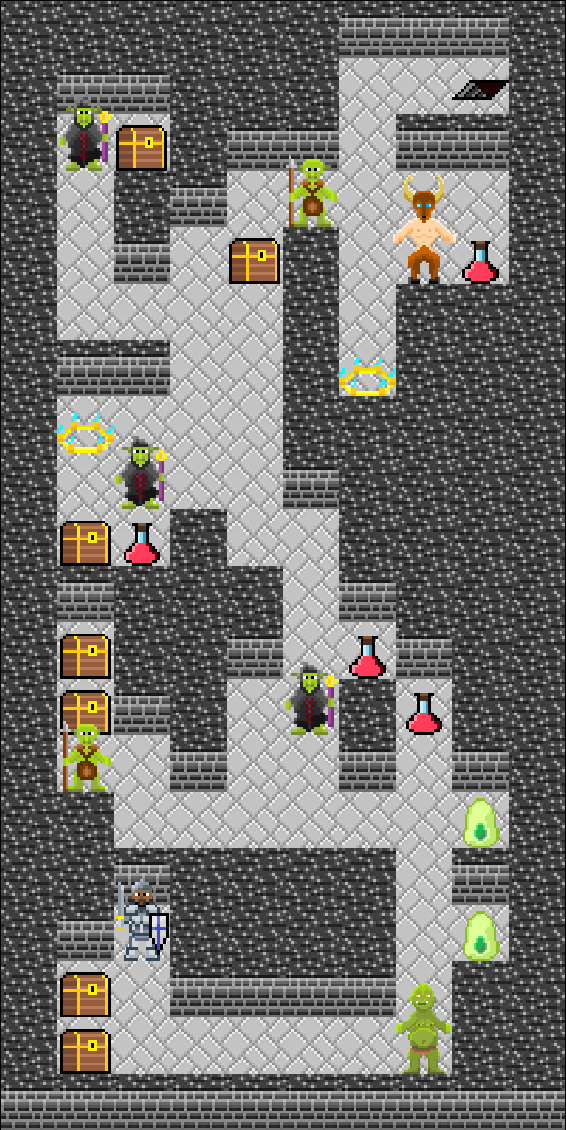}}~
\subfloat[Map 7]{\includegraphics[width=40px]{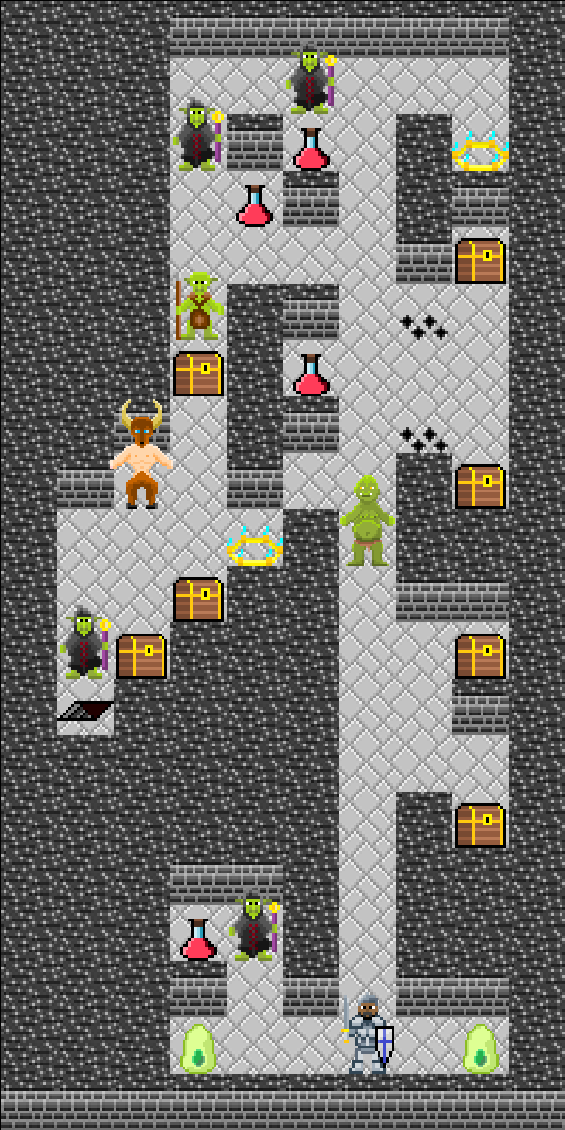}}~
\subfloat[Map 8]{\includegraphics[width=40px]{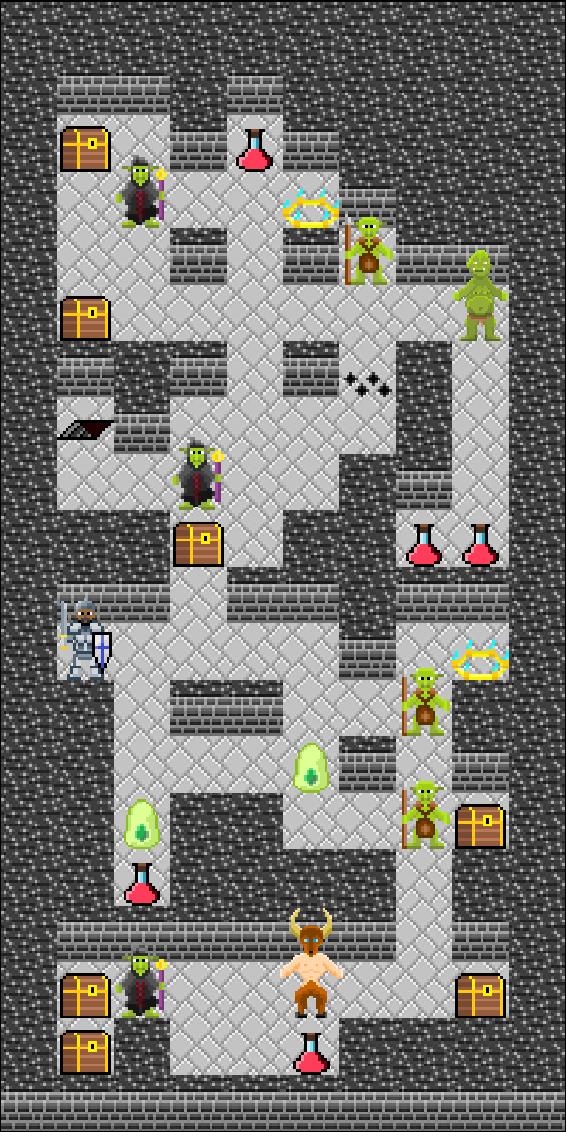}}~
\subfloat[Map 9]{\includegraphics[width=40px]{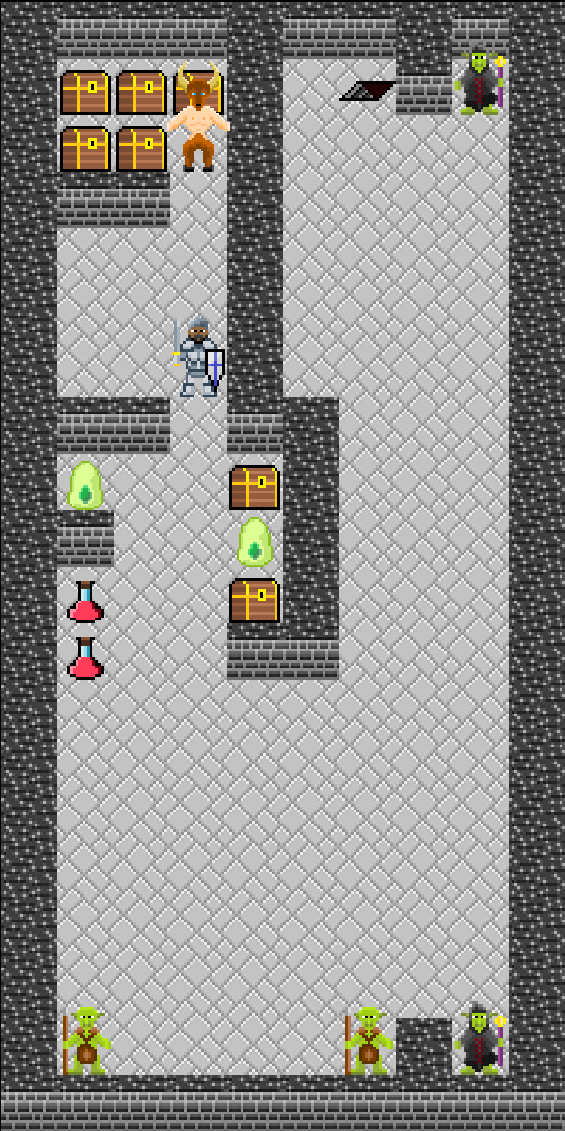}}~
\subfloat[Map 10]{\includegraphics[width=40px]{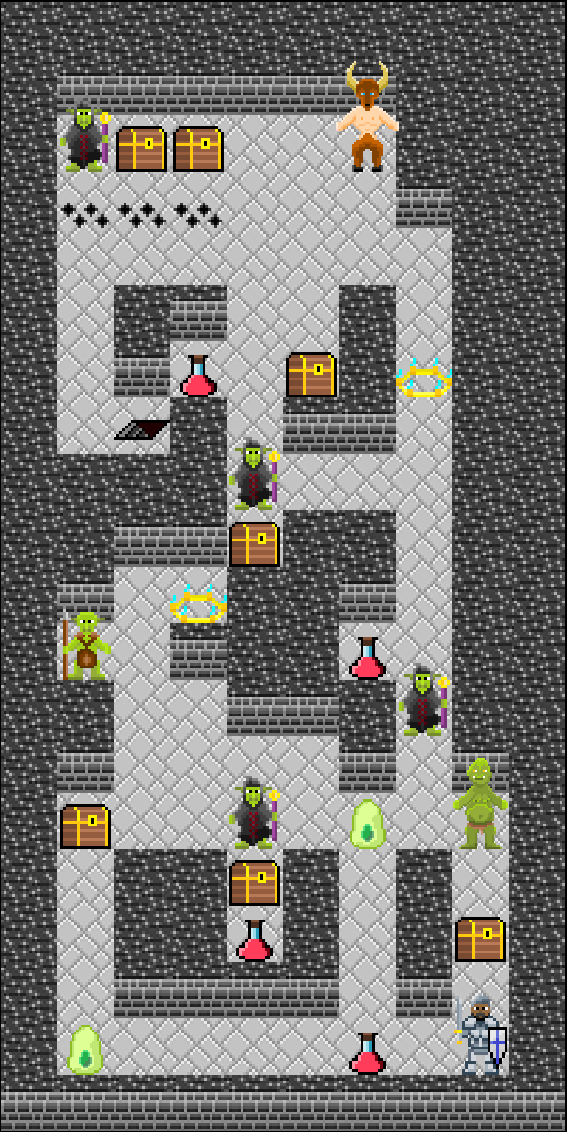}}~
\subfloat[Map 11]{\includegraphics[width=40px]{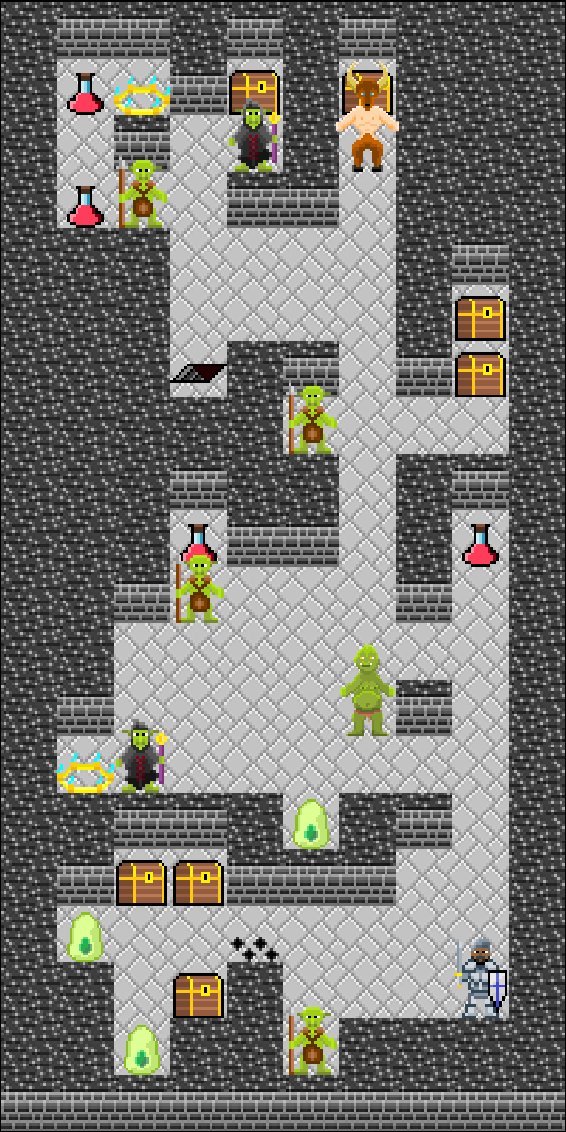}}
\caption{All 11 maps in the MiniDungeons 2 game.}
\label{fig:maps}
\end{figure*}

Since MiniDungeons 2 maps can combine interactive tiles and monsters in many different ways, the fitness of each individual is based on their utility in different maps. Evolving agents are tested on maps 1, 2, 3, 4, 7, and 10 of Fig.~\ref{fig:maps}: these maps capture many different playstyle patterns of MiniDungeons 2. The overall fitness of a chromosome is calculated by averaging its fitness across all playthroughs in these six maps. This averaged fitness score was then used to select genes (i.e. UCB1 replacement functions) for recombination, mutation and replacement.

The initial population of 100 individuals is created via the initialization process described above. Evolution uses an islands model \cite{whitley1999island} with 5 islands. Migration occurs in every generation. After migration, the five fittest individuals of each island are selected and placed into that island's \emph{mating pool}. Based on preliminary experiments, elitism was set to 15\% of the population. Before crossover, all individuals in the mating pool have a 10\% chance of mutating. Mutation replaces the chromosome with a random chromosome via the same initialization process described above. After mutation, the mating pool undergoes \emph{crossover}: two random chromosomes from the mating pool are crossed-over to create two offspring, formed by exchanging randomly selected sub-trees between the parents. During crossover, every individual in the mating pool has an equal chance to be selected, regardless of fitness. After this process is repeated for every island, a new population is generated and evaluated for fitness.  

\section{Experiments}\label{sec:experiments}

The purpose of evolving UCB1 replacement functions is to optimize the agents' behavior relative to the persona-defining utility function. Below, we describe the results of evolving the agents, comparing them with the standard UCB1 function and using the evolved agents to playtest maps.

\subsection{Experimental Protocol}\label{sec:protocol}

For the purposes of evolving the four agents' tree policy equations, six maps are played by the agent and the fitness score is calculated as described in Section \ref{sec:evolving_method}. However, for evaluating the agents' performance a broader set of maps is used: all 11 maps of Fig.~\ref{fig:maps} are tested in Sections \ref{sec:persona_analysis} and \ref{sec:level_personas}. 
To cater for the stochastic nature of MCTS, each map is simulated in 50 trials by each agent. To cater for the stochastic nature of evolutionary algorithms, 3 independent evolutionary runs of 100 generations are performed with a population of 100 individuals, following the process described in \ref{sec:evolving_method}. The best performing run (based on the persona's core priority, e.g. monsters killed for the Monster Killer) is chosen among the three evolutionary runs and reported here.

For the purposes of assessing the performance of evolved personas, \emph{baseline MCTS agents} using the UCB1 tree policy of Eq.~\ref{eq:ucb} are used to simulate the 11 maps in 50 trials each. Each baseline persona uses UCB1 for its tree policy but then backpropagates the persona-specific utility of Eqs.~(\ref{eq:baseline_r}-\ref{eq:baseline_c}) after each simulation. All reported significance testing is performed through Student's two-tailed $t$-tests, assuming unequal variance, with an error of $5\%$; when comparing between maps, the 50 playthroughs of each persona are tested for significance. Otherwise, 95\% confidence intervals are calculated via the standard deviation of all playthroughs in all maps.

\subsection{Persona Evolution}\label{sec:evolution}

For all personas, fitness starts converging after approximately 20 generations and from then on improves only marginally.
The best evolved tree policy equation for each persona in its raw form often has duplicate variables and can be simplified. We simplified each of the fittest equations at the end of 100 generations and list them as eq.~(\ref{eq:ucb_r}--\ref{eq:ucb_c}). Some of these equations are quite convoluted but some interesting patterns can be gleaned. The tree policy for the Runner in eq.~\eqref{eq:ucb_r} strongly prioritizes the proximity to the exit variable but also has a negative factor for health left (i.e. it actively prefers reaching the exit with low health). The tree policy for the Monster Killer in eq.~\eqref{eq:ucb_mk} aggressively prioritizes monsters slain and proximity to the exit as these variables are multiplied to any other metric; interestingly this tree policy is the only one among eq.~(\ref{eq:ucb_r}--\ref{eq:ucb_c}) which does not include $R$ (the average reward). The tree policy for the Treasure Collector in eq.~\eqref{eq:ucb_tc} is the only linear equation (a weighted sum) which includes an added 0.19 constant which obviously does not affect the tree policy as it is added to all possible moves. More interestingly, the Treasure Collector policy puts more weight on potions drunk and monsters slain (2 for each) rather than on treasures opened (weight of 1). The Completionist has the most complex tree policy in eq.~\eqref{eq:ucb_c}. It puts a surprising emphasis on steps taken (multiplied to most components), monsters slain and proximity to exit (despite the fact that it also subtracts $PE$); most surprisingly, it only includes treasures opened ($TO$) once and with a negative weight, meaning that it actively tries to reduce the number of treasures opened despite the fact that the utility (and fitness) of eq.~\eqref{eq:baseline_c} actively rewards treasures as a member of the interactive objects set.

\begin{align}
t_R =& 6.235{\cdot}ST{\cdot}PE^2{\cdot}(PE+1)+R{\cdot}(1-HL)
\label{eq:ucb_r}\\
t_{MK} =& 4{\cdot}MS{\cdot}PE{\cdot}(MS+2{\cdot}HL{\cdot}(PE-IC))
\label{eq:ucb_mk}\\
t_{TC} =& 2{\cdot}PD+2{\cdot}MS+TO\nonumber \\
&+3{\cdot}R+ST+PE+0.19
\label{eq:ucb_tc}\\
t_C =&
%
ST{\cdot}MS{\cdot}(ST^2{\cdot}MS + IC) 
+R-TO+IC
\nonumber \\
&-PE
+2{\cdot}ST{\cdot}PE{\cdot}(ST{\cdot}MS{\cdot} + 1)
\label{eq:ucb_c}
\end{align}

\subsection{Comparing Personas}\label{sec:persona_analysis}

It is not sufficient to assess the evolved personas based on their fitness scores alone. This section tests the final best personas based on what types of content they interact with in all the maps of MiniDungeons 2. The focus is on comparisons with UCB1 MCTS personas (as baselines) regarding agents' efficiency in achieving their core priorities, but also on comparisons between different personas' play behavior.

\subsubsection{Overall Performance}\label{sec:averages}

\begin{table}
\centering
\caption{Average scores in several game metrics for evolved and baseline personas. Results are averaged from 50 independent playthroughs of the best personas in each of the 11 maps. Results include the 95\% confidence interval.}
\label{table:averages}
\begin{tabular}{|l| r@{}l@{\hspace{1em}} r@{}l@{\hspace{1em}} r@{}l@{\hspace{1em}} r@{}l|}
\hline
Metric & \multicolumn{2}{c}{R} & \multicolumn{2}{c}{MK} & \multicolumn{2}{c}{TC} & \multicolumn{2}{c|}{C} \\
\hline 
\multicolumn{9}{|l|}{Evolved}\\
\hline 
Monsters &	$46\%$&${\pm}2\%$	&	$67\%$&${\pm}2\%$	&	$59\%$&${\pm}2\%$	&	$66\%$&${\pm}3\%$	\\
Potions &	$8\%$&${\pm}1\%$	&	$4\%$&${\pm}1\%$	&	$25\%$&${\pm}2\%$	&	$8\%$&${\pm}1\%$	\\
Treasures &	$10\%$&${\pm}1\%$	&	$7\%$&${\pm}1\%$	&	$35\%$&${\pm}2\%$	&	$10\%$&${\pm}1\%$	\\
Interactive Objects &	$25\%$&${\pm}1\%$	&	$33\%$&${\pm}1\%$	&	$41\%$&${\pm}1\%$	&	$34\%$&${\pm}1\%$	\\
\hline
Win Rate &	$100\%$&${\pm}0\%$	&	$73\%$&${\pm}4\%$	&	$54\%$&${\pm}4\%$	&	$100\%$&${\pm}0\%$	\\
Time (sec) &	$3.2$&${\pm}0.3$	&	$83$&${\pm}11$	&	$151$&${\pm}12$	&	$8.1$&${\pm}1.2$	\\
\hline
\multicolumn{9}{|l|}{Baseline}\\
\hline
Monsters &	$25\%$&${\pm}1\%$	&	$29\%$&${\pm}1\%$	&	$25\%$&${\pm}1\%$	&	$28\%$&${\pm}0\%$	\\
Potions &	$5\%$&${\pm}1\%$	&	$5\%$&${\pm}1\%$	&	$6\%$&${\pm}1\%$	&	$5\%$&${\pm}0\%$	\\
Treasures &	$5\%$&${\pm}1\%$	&	$5\%$&${\pm}1\%$	&	$17\%$&${\pm}2\%$	&	$6\%$&${\pm}0\%$	\\
Interactive Objects &	$13\%$&${\pm}1\%$	&	$16\%$&${\pm}1\%$	&	$19\%$&${\pm}1\%$	&	$15\%$&${\pm}0\%$	\\
\hline
Win Rate &	$10\%$&${\pm}3\%$	&	$12\%$&${\pm}3\%$	&	$9\%$&${\pm}2\%$	&	$13\%$&${\pm}0\%$	\\
Time (sec) &	$277$&${\pm}6$	&	$285$&${\pm}4$	&	$279$&${\pm}5$	&	$278$&${\pm}0$	\\
\hline
\end{tabular}
\end{table}

Table \ref{table:averages} shows the ratio of game objects each agent has interacted with (i.e. monsters, potions, treasures) on average in the 11 testbed maps of MiniDungeons 2. At a high level, the Monster Killer kills more monsters on average than the other personas, while the Treasure Collector collects more treasure and drinks more potions. A more detailed analysis in Section \ref{sec:ttest_personas} will shed more light on the differences between personas, as there are substantial deviations between maps.  Table \ref{table:averages} includes the win rate of different personas (i.e. instances where the agent reached the exit), as well as computation time to find a path (up to a maximum of 300 seconds). The evolved Runner persona is consistently able to reach the exit in all maps, and does so in far fewer steps and requiring far less computational time than all other personas, both evolved and baseline MCTS. Surprisingly, the evolved Completionist persona also completes all maps in all trials, despite the fact that it prioritizes interacting with as many game objects as possible; perhaps due to the latter strategy its computational time is double that of the evolved Runner. Since the Treasure Collector does not receive a large reward for finishing the level, it tends to roam around the map attempting to collect all treasures, and does not finish before a maximum allocated time in 46\% of all trials. The Monster Killer has a similarly low reward for finishing the level, however it only roams around certain maps until the allocated time runs out (27\% of all trials). It should be noted that the computation time is averaged from all playthroughs regardless of whether the agent reached the exit: if only won playthroughs are considered, the average computation time of the evolved Monster Killer (2.06 sec) is the lowest of all other evolved personas, while that of the evolved Treasure Collector remains high (at 23 sec). That said, considering only computation time of won games these values are still better than those of the baseline MCTS Runner (71 sec), Monster Killer (167 sec) and Completionist (133 sec); only the baseline Treasure Collector is relatively close (72 sec) to its evolved counterpart but it only wins in one map (map 8). Comparisons between baseline and evolved agents in the remaining game metrics of Table \ref{table:averages} will be detailed in Section \ref{sec:ttest_baselines}.

\subsubsection{Differences from the Baseline}\label{sec:ttest_baselines}

For the purposes of comparisons between evolved and baseline personas, the results of Table \ref{table:averages} are too noisy due to the sensitivity of persona behavior in different maps of MiniDungeons 2. For a more thorough comparison, Table \ref{table:ttest_baselines} compares the number of maps (out of 11) in which the different metrics are significantly higher for the evolved persona than the baseline persona of the same type (E), or significantly lower (B). Significance is tested via two-tailed Student's $t$-tests ($p<5\%$) comparing 50 playthroughs of each map per persona (evolved or baseline). 

Table \ref{table:ttest_baselines} shows that evolved personas score significantly higher (or lower, for computation time) in the different metrics in more maps than their baseline counterparts. A notable exception is the treasure ratio for the Runner, Monster Killer, and Completionist, as the MCTS personas collect more treasure in a comparable number of maps; however, these baseline MCTS agents do not complete the level in far more cases. Especially regarding win rates, the evolved personas are always superior (or at least not inferior) in all maps and for all personas. In comparison, baseline personas need more computational time and do not finish a level far more often as shown by their win rates in Table \ref{table:averages} (less than 15\% for all personas).
%

\begin{table}
\centering
\caption{Maps in which the shown metrics are significantly higher for the evolved persona (E) than the baseline persona of the same type, and maps in which the reverse is true (B).}
\label{table:ttest_baselines}
\begin{tabular}{|l|c@{\hspace{1em}}c || c@{\hspace{1em}}c || c@{\hspace{1em}}c || c@{\hspace{1em}}c |}
\hline
 & \multicolumn{2}{|c||}{R} & \multicolumn{2}{c||}{MK} & \multicolumn{2}{c||}{TC} & \multicolumn{2}{c|}{C} \\
\cline{2-9}
Metric & E & B & E & B & E & B & E & B \\
\hline
Monster Ratio	 & 8 & 1	 & 10 & 0	 & 10 & 1	 & 10 & 0	\\
Potion Ratio	 & 2 & 0	 & 2 & 1	 & 8 & 1	 & 2 & 0	\\
Treasure Ratio	 & 3 & 2	 & 2 & 5	 & 7 & 1	 & 3 & 3	\\
Interactive Objects Ratio	 & 9 & 1	 & 10 & 0	 & 10 & 1	 & 10 & 0	\\
\hline
Time (sec)	 & 0 & 11	 & 0 & 11	 & 0 & 10	 & 0 & 11	\\
Win Rate	 & 10 & 0	 & 7 & 0	 & 7 & 0	 & 10 & 0	\\
\hline
\end{tabular}
\end{table}

\subsubsection{Differences among Personas}\label{sec:ttest_personas}
Due to the large differences between MiniDungeons 2 maps in the different metrics of Table \ref{table:averages}, to compare whether (and how) the procedural personas play the game differently we evaluate the number of maps in which one persona has a significantly higher value for one metric than another persona. This comparison is summarized in Table \ref{table:ttest_personas}; significant differences are established from a $t$-test ($p<5\%$) between 50 playthroughs of each persona in one map. 
We are interested in seeing whether the evolved personas, which have been shown to be more efficient and robust at gameplaying, still maintain differentiation in those game metrics that make them unique (e.g. a Monster Killer persona should kill more monsters than other personas).

Analyzing the general differences between the evolved Monster Killer and other evolved personas in Table \ref{table:ttest_personas}, we see that its killed monsters ratio is higher for this persona; no other persona has a higher ratio in any map. The evolved Treasure Collector collects significantly more treasure in most maps (8 or 9 out of 11); the baseline Treasure Collector is close but is not superior from other baseline personas in as many maps. Interestingly, the evolved Completionist is underperforming in all relevant metrics compared to the Treasure Collector: it interacts with more game objects only in 1 map (the Treasure Collector has more interactive objects in 8 maps) and generally drinks fewer potions and collects less treasure. 

It is therefore obvious from Table \ref{table:ttest_personas} that the Completionist is inferior to the Treasure Collector apart from the fact that it kills more monsters in two maps. This is surprising due to the fact that this persona explicitly rewarded a high interactive objects ratio in the fitness for deciding its tree policy, and when scoring the default policy. On the other hand, the evolved Completionist persona is the only persona apart from the Runner which wins all 50 playthroughs in all 11 maps while still interacting with more game objects (primarily monsters) than the Runner. Even the baseline Completionist persona has a high win rate compared to the baseline Runner (see Table \ref{table:averages} while other baseline personas interact with more objects. It is our assumption that using the interactive objects ratio for the utility score (and one would assume as a fitness) creates an imbalance between interacting with an object and approaching the exit.
For instance, most maps have 5 to 8 treasure tiles and thus a Treasure Collector would have a higher utility gain by collecting a couple during a playthrough rather than by getting a few steps closer to the exit; instead, when maps have around 20 interactive objects then a completionist interacting with a couple of them will have a lower utility gain than approaching the exit. Completionist agents thus favor reaching the exit, although not as aggressively as the the Runner as there is some reward (however slight) for deviating from the path.


\begin{table}
\centering
\caption{Maps in which the shown metrics are significantly higher for the persona on the row than in the persona in the column. }
\label{table:ttest_personas}
\begin{tabular}{|l| c c c c || l | c c c c|}
\hline & \multicolumn{4}{c||}{Evolved} & & \multicolumn{4}{c|}{Baseline}\\ \cline{2-10} 
& R & MK & TC & C & & R & MK & TC & C\\		
\hline \multicolumn{10}{|l|}{Monster Ratio}\\ \hline 									
R &	--- &	0 &	3 &	1 &	R &	--- &	0 &	1 &	0 \\
MK &	\textbf{8} & 	--- &	\textbf{6} &	\textbf{3} &	MK &	\textbf{5} & 	--- &	\textbf{5} &	\textbf{2} \\
TC &	7 & 	3 &	--- &	3 &	TC &	2 & 	1 &	--- &	1 \\
C &	8 &	2 &	6 &	--- &	C &	3 &	0 &	2 &	--- \\
							
\hline \multicolumn{10}{|l|}{Potion Ratio}\\ \hline 									
R &	--- &	2 &	1 &	0 &	R &	--- &	0 &	1 &	0 \\
MK &	1 & 	--- &	0 &	1 &	MK &	0 & 	--- &	1 &	0 \\
TC &	8 & 	8 &	--- &	8 &	TC &	3 & 	3 &	--- &	3 \\
C &	0 &	2 &	1 &	--- &	C &	0 &	0 &	1 &	--- \\
									
\hline \multicolumn{10}{|l|}{Treasure Ratio}\\ \hline									
R &	--- &	3 &	0 &	0 &	R &	--- &	1 &	0 &	1 \\
MK &	0 & 	--- &	0 &	0 &	MK &	2 & 	--- &	0 &	0 \\
TC &	\textbf{9} & 	\textbf{9} &	--- &	\textbf{8} &	TC &	\textbf{6} & 	\textbf{7} &	--- &	\textbf{6} \\
C &	2 &	3 &	0 &	--- &	C &	1 &	0 &	0 &	--- \\
		
\hline \multicolumn{10}{|l|}{Interactive Object Ratio}\\ \hline 									
R &	--- &	3 &	1 &	1 &	R &	--- &	0 &	0 &	0 \\
MK &	6 & 	--- &	1 &	2 &	MK &	4 & 	--- &	0 &	1 \\
TC &	9 & 	8 &	--- &	8 &	TC &	6 & 	5 &	--- &	6 \\
C &	\textbf{8} &	\textbf{4} &	\textbf{1} &	--- &	C &	\textbf{2} &	\textbf{0} &	\textbf{0} &	--- \\ 
        
\hline \multicolumn{10}{|l|}{Time}\\ \hline 									
R &	--- &	2 &	0 &	1 &	R &	--- &	0 &	0 &	2 \\
MK &	3 & 	--- &	1 &	3 &	MK &	1 & 	--- &	1 &	3 \\
TC &	11 & 	10 &	--- &	10 &	TC &	2 & 	1 &	--- &	2 \\
C &	6 &	4 &	0 &	--- &	C &	1 &	0 &	1 &	--- \\
									
\hline \multicolumn{10}{|l|}{Win Rate}\\ \hline 									
R &	--- &	3 &	7 &	0 &	R &	--- &	0 &	1 &	0 \\
MK &	0 & 	--- &	4 &	0 &	MK &	1 & 	--- &	1 &	0 \\
TC &	0 & 	1 &	--- &	0 &	TC &	0 & 	0 &	--- &	0 \\
C &	0 &	3 &	7 &	--- &	C &	1 &	1 &	1 &	--- \\
\hline
\end{tabular}
\end{table}

\subsection{Evaluating Levels with Personas}\label{sec:level_personas}

Procedural personas can be used for many different purposes, such as modeling players based on the similarity of players' actions with a persona's action \cite{holmgard2015evolving}. However, personas can also be used to evaluate game levels by creating artificial playtraces; this can be used as feedback to a human designer when procedural personas test an authored level, but also as a way to improve computer generated levels in a search-based approach driven by the artificial playtraces of personas in similuations \cite{liapis2015personacritics}. In this paper, the former approach is followed and we evaluate which patterns of game levels affect the performance of different personas. We will only use the evolved personas, as they are overall superior.

\begin{figure}[t]
\centering
\includegraphics[width=0.95\columnwidth]{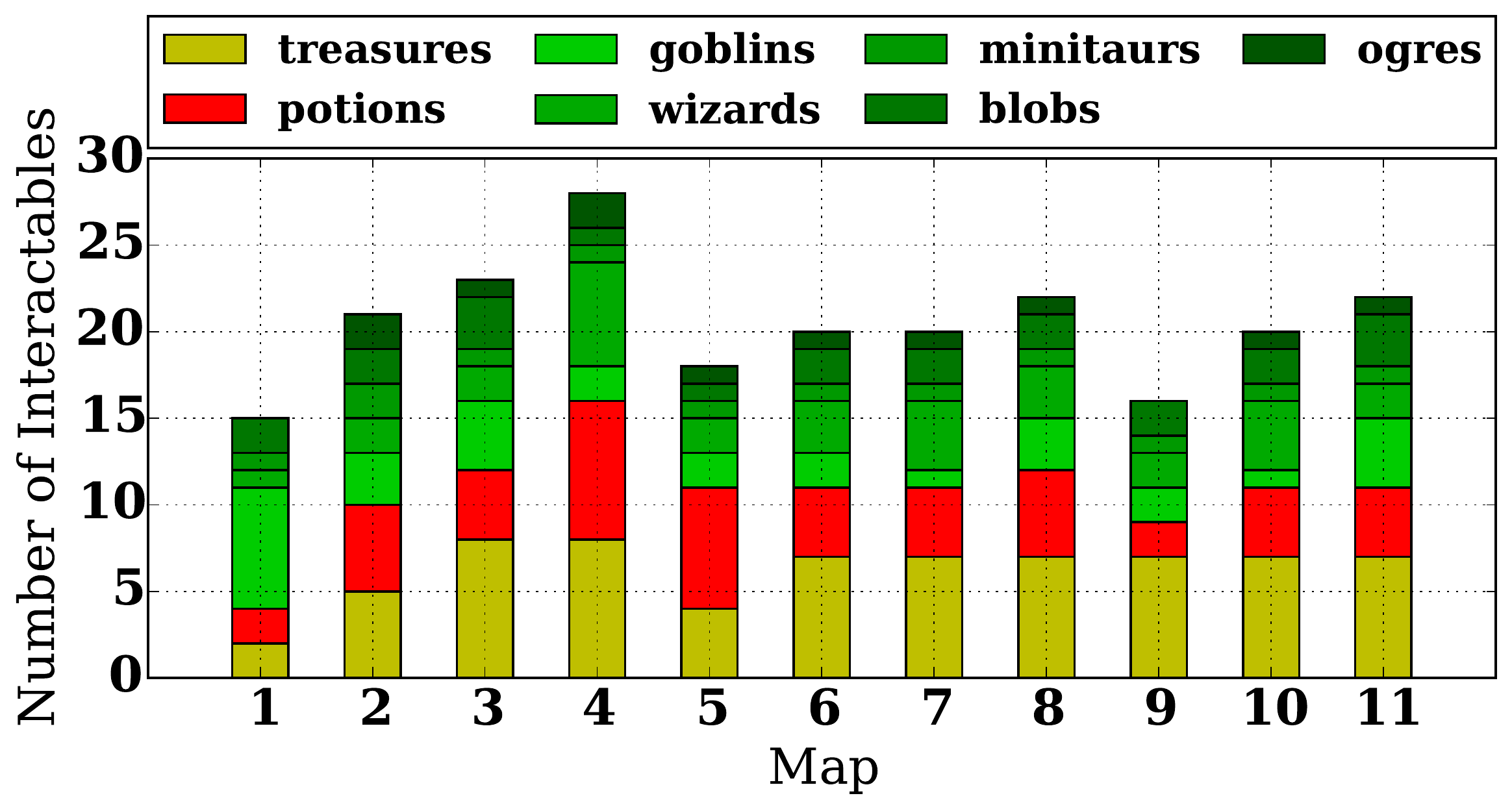}
\caption{Number of interactive objects in MD2 maps.}
\label{fig:stacked}
\end{figure}

In order to first identify what the differences are between levels, Fig.~\ref{fig:stacked} shows the number of interactive objects (i.e. potions, treasure, and the different types of monsters) contained in each map used in this analysis. Obviously, some maps have fewer interactive objects (map 1, map 9), and some maps have more potions and few treasures (e.g. map 5) or vice versa (e.g. map 9). There are also many differences in the types of monsters favored; while all maps include at least one minitaur (map 2 has two of them), some maps do not include ogres (map 1, map 9) and some maps have more ranged goblin enemies than melee goblin enemies (e.g. map  4, map 10) or vice versa (e.g. map 1). It should be noted that besides interactive objects, these maps differ in terms of other types of tiles, e.g. seven maps contain a set of portals allowing shortcuts through the level, while six maps contain one or more traps which deal damage when the tile is visited.

A broad range of metrics on the levels' structure alone (before simulation) were collected from the 11 maps of {MiniDungeons 2}. These include the number of interactive objects of Fig.~\ref{fig:stacked}, the number of portals and traps, the number of wall tiles, choke points and dead ends (tiles with only two or one connected passable tiles, respectively), length of the shortest path between entrance and exit and many others. The metrics of all maps were then analyzed in terms of their correlations with the performance of each persona in the same map. For the sake of brevity, only correlations with each persona's win rate and \emph{core priority} will be discussed: i.e. for the Runner computation time is the core priority, treasure ratio for Treasure Collector, monster ratio for Monster Killer and interactive object ratio for Completionist. While many of the level metrics were found to be correlated with these persona metrics, due the small sample size (11 maps) only a handful of significant correlations were found ($p<0.05$ of the Pearson's correlation coefficient, which is also reported as $r$) . 

For the Runner, computation time was significantly correlated with the length of the shortest path between entrance and exit ($r=0.71$). This is not surprising, since the Runner is efficient at finding a short route to the exit and thus requires less computation time if the exit is nearby. For the Treasure Collector, the ratio of collected treasures has a significant negative correlation with the number of walls ($r=-0.69$) and a significant positive correlation with the number of open areas  ($r=0.63$); open areas are tiles where all adjacent tiles are non-walls, thus it is not surprising that walls and open areas have opposite effects. It seems that the Treasure Collector is less able to handle winding corridors. The win ratio of the Treasure Collector persona has a significant negative correlation with  the shortest path length to the exit ($r=-0.62$), again pointing at this persona's poor performance in winding maze-like corridors. While no significant correlations were found for the Monster Killer, the Completionist's interactive object ratio is negatively correlated with the number of treasures in the map ($r=-0.65$). This is not surprising as the Completionist's evolved tree policy in eq.~\eqref{eq:ucb_c} actively discourages opening treasure chests, so the more of those there are in the map the fewer the Completionist's interactions with game objects.

\begin{figure}[t]
\centering
\subfloat[Map 6]{\includegraphics[width=0.45\columnwidth]{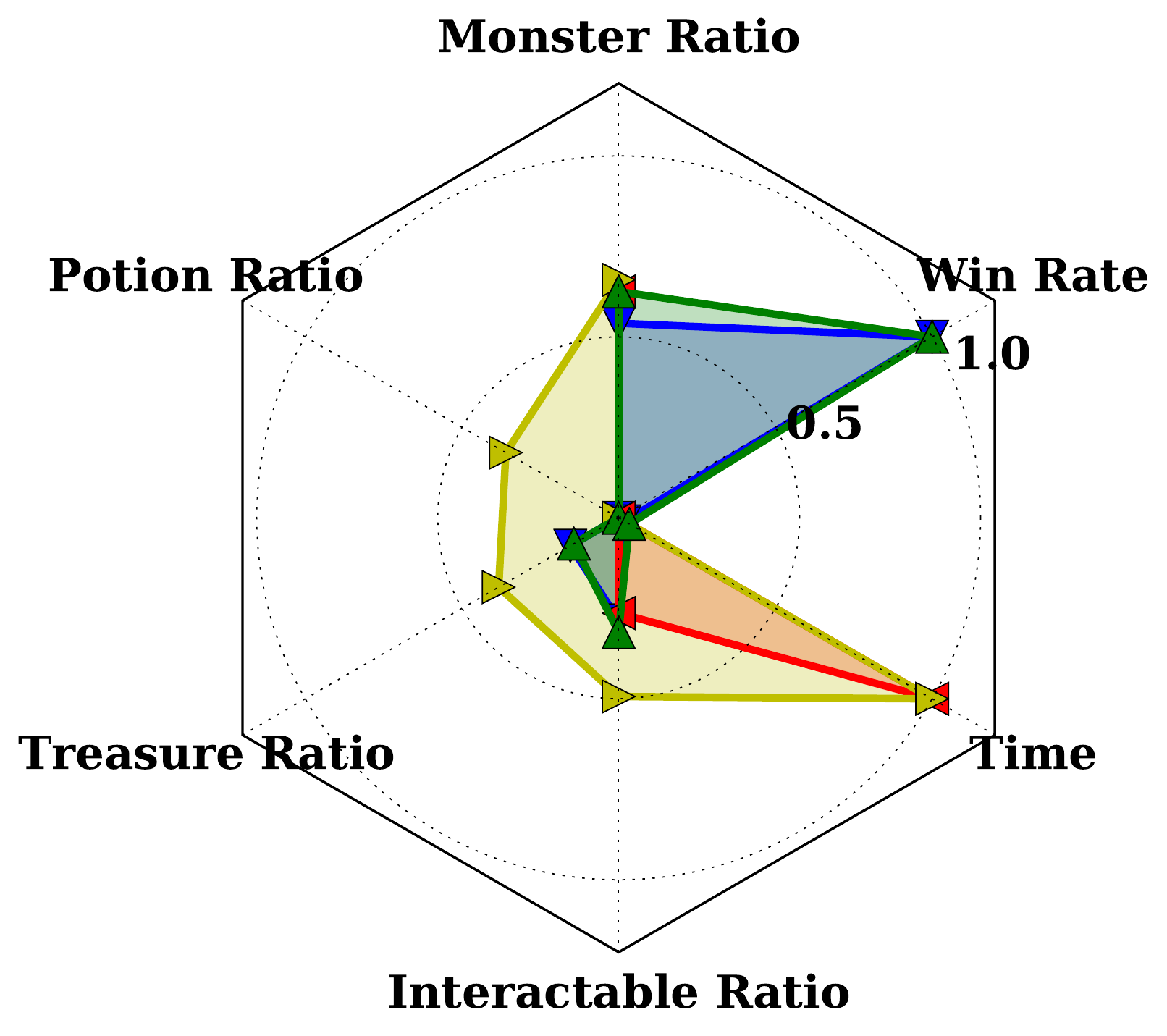}}
\subfloat[Map 8]{\includegraphics[width=0.45\columnwidth]{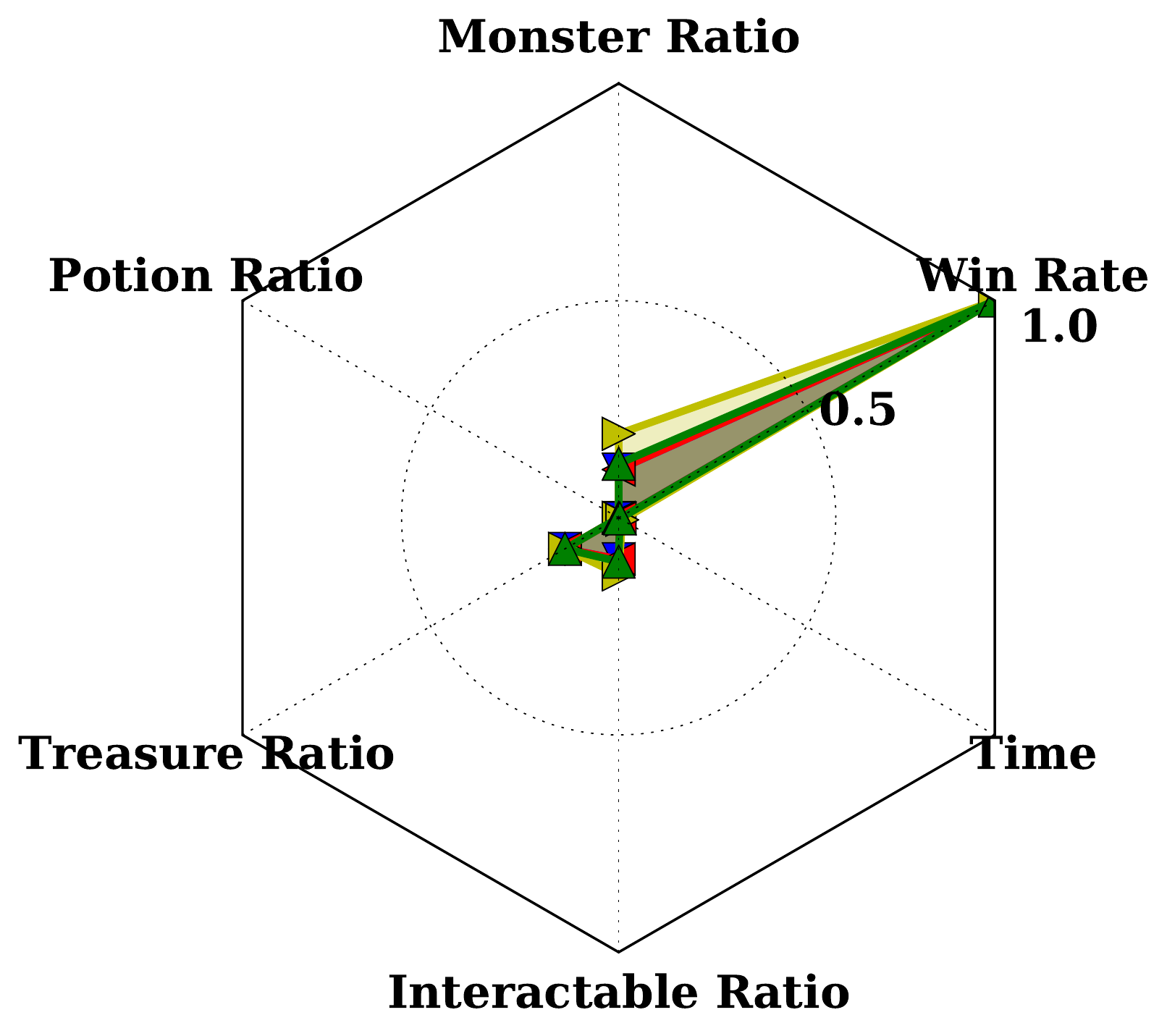}}\\
\subfloat{\includegraphics[width=\columnwidth]{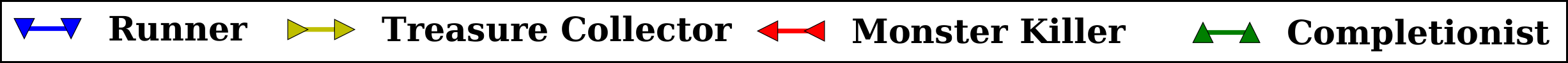}}
\caption{Metrics of different personas in the same map.}
\label{fig:radars}
\end{figure}

In order to see how the maps' layout can affect the diversity of playthroughs among personas, we choose two indicative maps to analyze; map 6 which has the most (significant) differences in all possible pairs of personas and for all metrics of Table \ref{table:ttest_personas}, and map 8 which has the fewest differences. The values of these metrics for different personas in each map are shown in Fig.~\ref{fig:radars}, averaged from 50 playthroughs.

\begin{figure}[t]
\centering
\subfloat[R]{\includegraphics[trim={285px 50px 285px 0},clip,width=0.23\columnwidth]{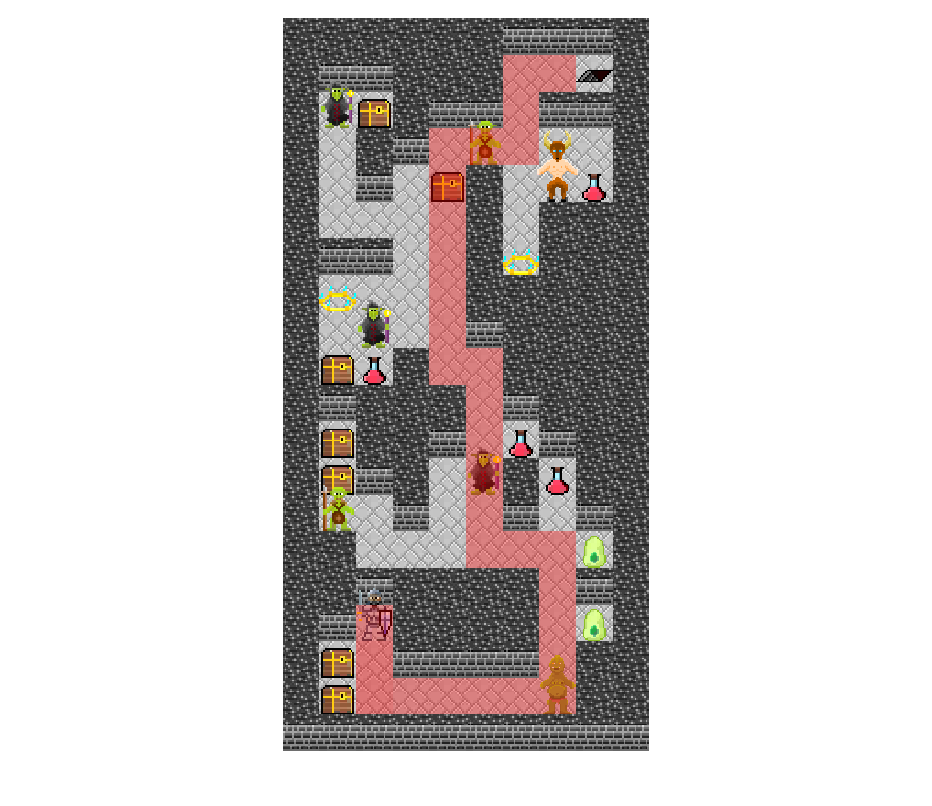}}~
\subfloat[MK]{\includegraphics[trim={285px 50px 285px 0},clip,width=0.23\columnwidth]{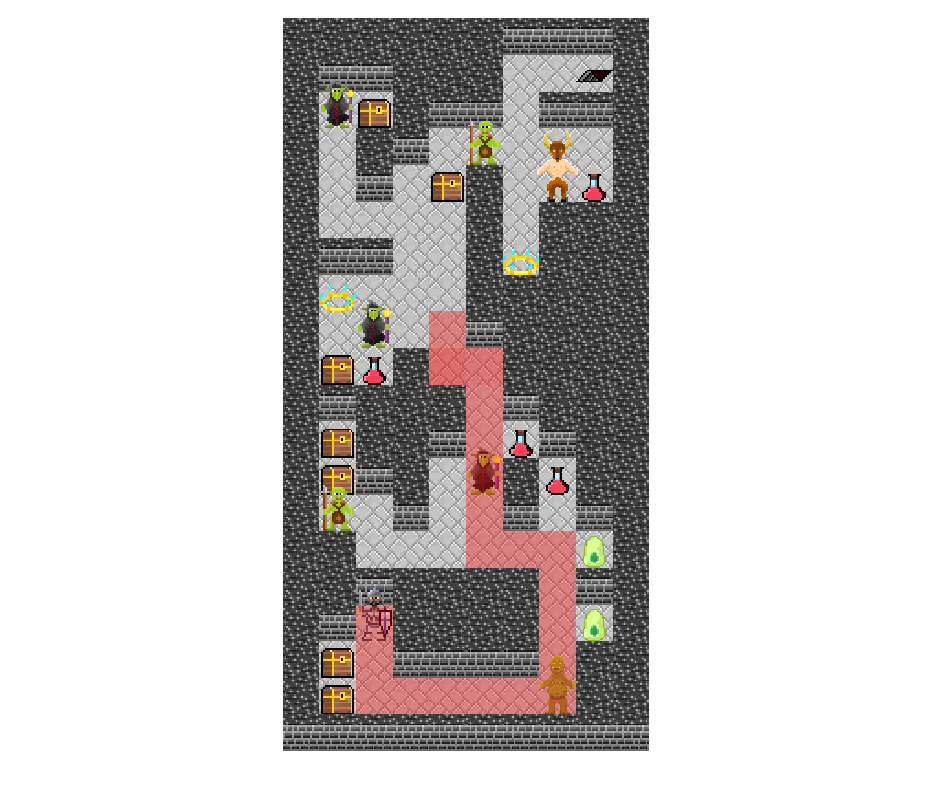}\label{fig:heatmap_map6_mk}}~
\subfloat[TC]{\includegraphics[trim={285px 50px 285px 0},clip,width=0.23\columnwidth]{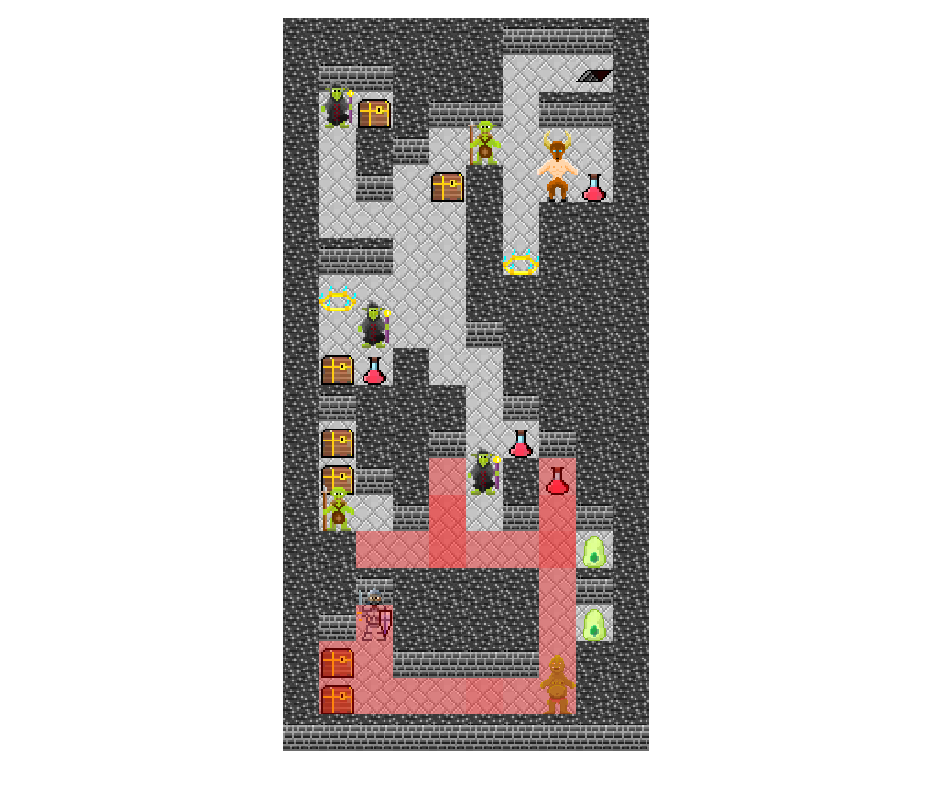}}~
\subfloat[C]{\includegraphics[trim={285px 50px 285px 0},clip,width=0.23\columnwidth]{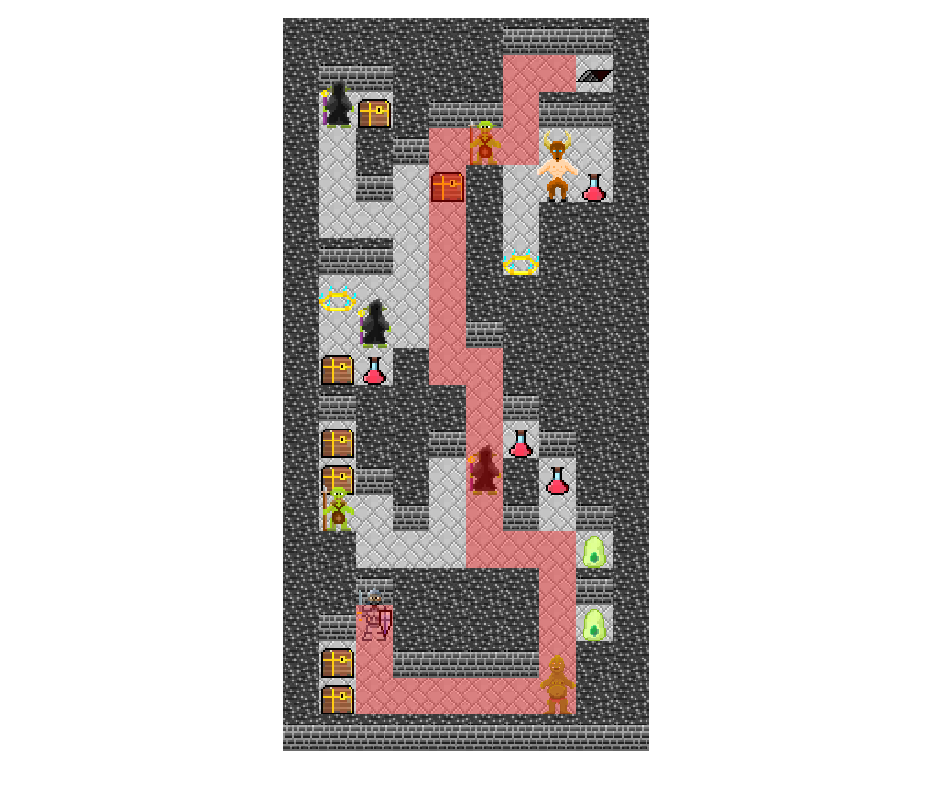}}
\caption{Heatmaps of persona behavior in map 6.}
\label{fig:heatmap_map6}
\end{figure}

In map 6, the Runner and Completionist reach the exit in all playthroughs while the Monster Killer and Treasure Collector never reach the exit as their computation time consistently reaches the timeout limit of 300 sec. Surprisingly, the Treasure Collector kills more monsters (66\%), collects more treasure (38\%), drinks more potions (36\%) and generally interacts with more game objects (49\%) than all other personas. The Runner persona manages to collect more treasure (15\%) than both the Monster Killer (0\%) and the Completionist (14\%). The Completionist has the second highest interactive object ratio (32\%). Based on the heatmaps of Fig.~\ref{fig:heatmap_map6}, the Monster Killer and Treasure Collector are shown to roam around the map and then get blocked from taking a decision until the 300 sec timeout. The Runner and Completionist on the other hand follow a similar path to the exit (top right) which is actually the shortest path. Only the Treasure Collector gets the two unguarded treasures next to the entrance (bottom left), while the wounded Monster Killer (due to combat with the ogre and two blobs) ignores both unguarded potions along its path in Fig.~\ref{fig:heatmap_map6_mk}. Interestingly, no persona uses the portal.

\begin{figure}[t]
\centering
\subfloat[R]{\includegraphics[trim={285px 50px 285px 0},clip,width=0.23\columnwidth]{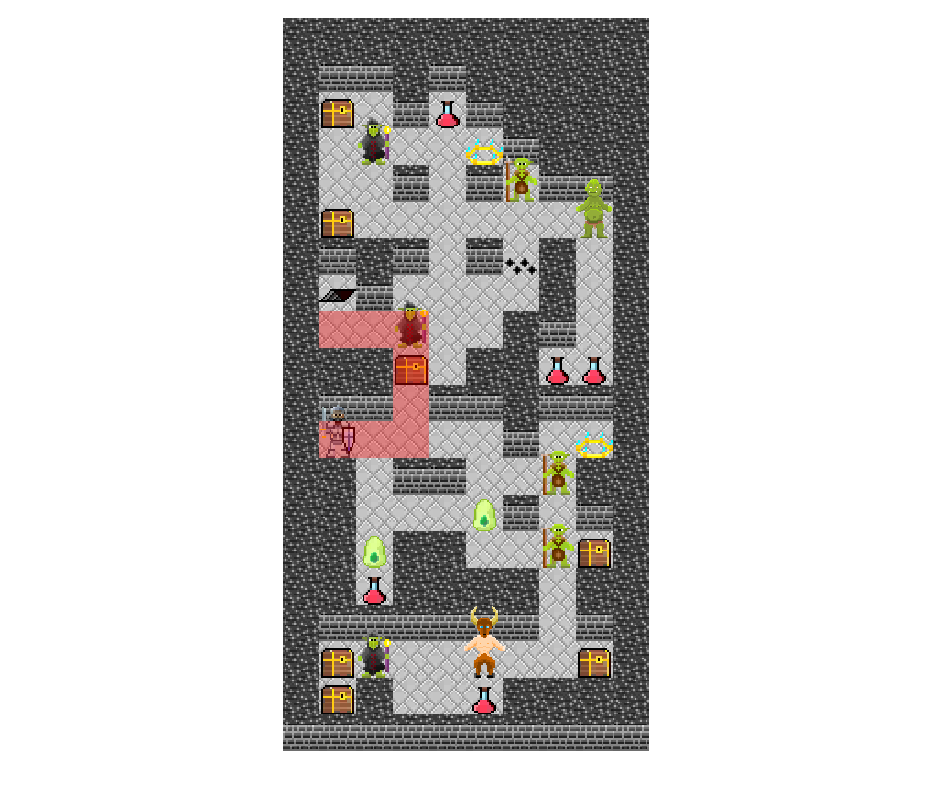}}~
\subfloat[MK]{\includegraphics[trim={285px 50px 285px 0},clip,width=0.23\columnwidth]{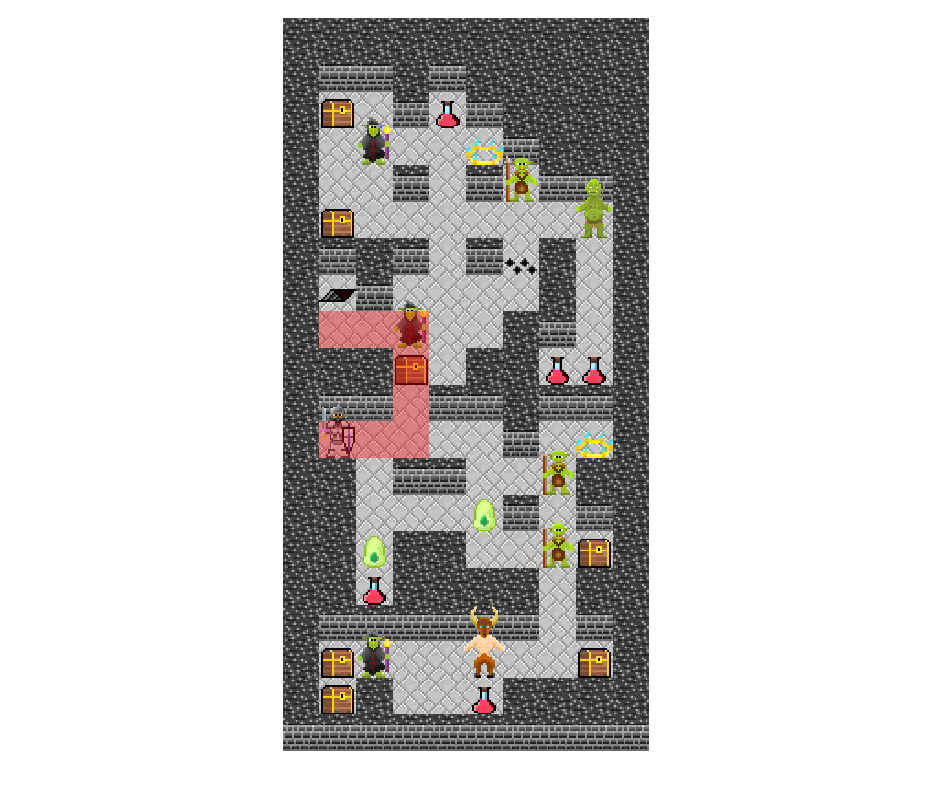}}~
\subfloat[TC]{\includegraphics[trim={285px 50px 285px 0},clip,width=0.23\columnwidth]{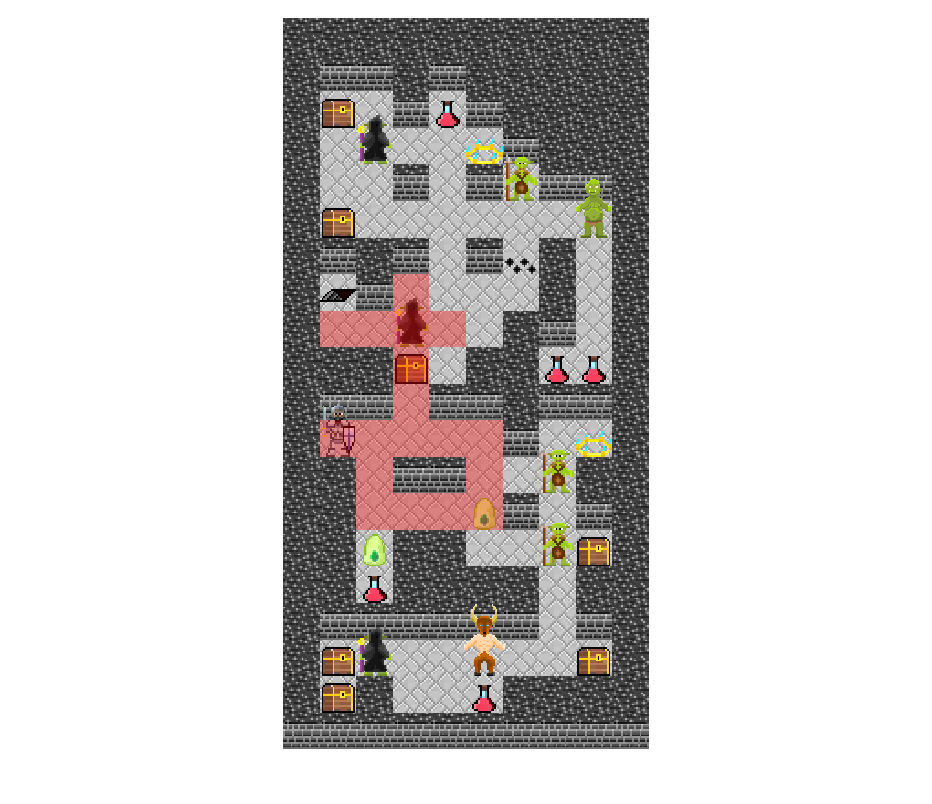}\label{fig:heatmap_map8_tc}}~
\subfloat[C]{\includegraphics[trim={285px 50px 285px 0},clip,width=0.23\columnwidth]{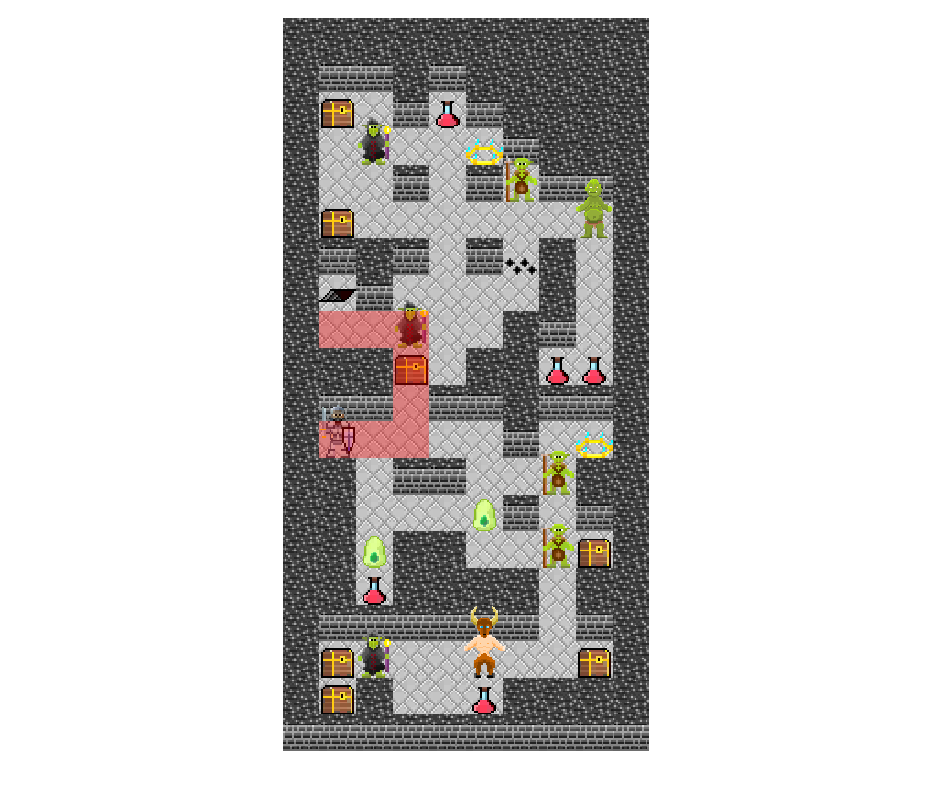}}
\caption{Heatmaps of persona behavior in map 8.}
\label{fig:heatmap_map8}
\end{figure}

In map 8, all personas reach the exit in all playthroughs with minimal computational time, and generally their other metrics are also similar. Again the monsters killed across 50 playthroughs are lower for the Monster Killer (11\%), than for the Treasure Collector (19\%) and slightly lower than the Completionist (12\%). All personas drink no potions and collect one treasure (which is mandatory in order to reach the exit as shown in Fig.~\ref{fig:heatmap_map8}); therefore the difference in interactive object ratio is solely due to more monsters killed by the Treasure Collector.
It is worth noting that Fig.~\ref{fig:heatmap_map8_tc} shows how the Treasure Collector may spend more time roaming around the map. On average, the Treasure Collector needs more computation time (2.8 sec) than the Monster Killer and Runner (0.8 sec for both). Indeed, the Treasure Collector takes on average 11.6 actions (the Runner takes 8, the Monster Killer 9 and the Completionist 9.2). This persona's behavior differs from playthrough to playthrough: in the one shown in Fig.~\ref{fig:heatmap_map8_tc}, the Treasure Collector took 23 actions and killed 5 monsters, which simply walked towards the agent (without the agent needing to explore the map).

Finally, it would be interesting to see if there are maps which are ``preferred'' by all personas. Using the priorities mentioned above (lowest computation time for Runner, highest treasure ratio for Treasure Collector, highest monster ratio for Monster Killer and highest interactive object ratio for Completionist), we find that the best map for the Monster Killer is map 9, in which it kills all monsters in all playthroughs, and the worst is map 8. Map 8 is also deemed the worst by the Completionist, while map 1 is deemed the best. In contrast, map 8 is deemed the best for the Runner, and map 1 the worst. For the Treasure Collector map 5 is the best and map 10 is the worst. It is telling that often the worst map for one persona is the best for another, pointing to the fact that different priorities combined with different behaviors to achieve those priorities can saturate how each persona assesses the maps.

\section{Discussion}\label{sec:discussion}

The experiments of Section \ref{sec:experiments} demonstrated that the evolved personas were able to play the game more efficiently ---requiring less computational time--- than the baseline UCB1 agents, and were more robust in completing each map by reaching the exit. Despite being efficient in completing most levels (or all levels in the case of the Runner and Completionist), the evolved personas still differentiate their playstyle and in most maps perform better than other personas with regard to their core priority.
The exception is the Completionist, which seems to be an inferior form of the Treasure Collector; however this persona has an important benefit in that it completes all the levels consistently while performing more interactions than the Runner.
Looking at the effect that each map of MiniDungeons 2 had on each persona, we identify that different personas are sensitive to different level patterns.
Such findings could influence the level design or game design of MiniDungeons 2 (i.e. creating more open areas and fewer winding corridors), or lead to a re-design of the fitness or utility functions e.g. as we find that treasures are not favored by Completionists.
Finally, in most cases results differ in terms of which map from the set is best or worst for each persona according to their priorities and playstyles. 

Since procedural personas are intended to be a design tool, they are inherently subjective in the sense that the utility functions should be constructed by a game or level designer interested in testing their content for the game. The experiments provided here support the persona concept as useful for fulfilling specific core priorities for a game of the scope and size of MiniDungeons 2. How the method scales to games of higher complexity is an open question; any game could in principle be tested using procedural personas, as long as it includes agent control methods that can be optimized towards a particular utility function. The specification of the utility function is a complex issue for the procedural persona method: the concept is useful from a design perspective only to the extent that game designers are capable of defining appropriate utility sources and ways of weighting them. One approach to solving this problem could be learning utility functions from demonstration: e.g. from groups of observed players or from designers playing in different styles to demonstrate what particular personas should play like. This could be enabled via methods such as inverse reinforcement learning, driven by evolution or other methods. Regardless, the proposed method is supported in general by the fact that the personas exhibit significantly different behaviors in the same environments, driven by simple utility functions; it is thus likely that game creators would be able to use the personas to inform their content creation process. It also suggests that personas could successfully be integrated as critics in a procedural level generation system; this was previously done for MiniDungeons, a simpler game with simpler persona implementations~\cite{liapis2015personacritics}.

Another direction for future work is to validate \emph{a posteriori} the ability of the defined personas and their behavior to map to real human playtraces. This could allow players to be mapped to one of the four personas based on the similarity of persona and player gameplay traces either on the action-by action-level or on a more macro-strategy level, as done in \cite{holmgard2015evolving}. However, the current experiments do not include human players as they test how our method can allow game/level designers to define archetypal personas \emph{a priori}, before even showing the game or level to players. The experiments have demonstrated that different behaviors can be encoded in such a way, and that the persona's behaviors (in terms e.g. of monsters killed) largely match the designers' stated intentions.

\section{Conclusion}
This paper presented the general concept of procedural personas, a framework for generative player modeling for automatic playtesting. Procedural personas represent a potentially general framework for representing archetypal playstyles, based on decision theory, that could inform game designers about properties of their game content as it is being created. The experiments reported in this paper show that personas are capable of showing different interaction patterns in response to game content and can help map out the playspace afforded by game levels as those are being designed. By combining evolution and MCTS, we produce a set of personas that show what different play styles might look like in MiniDungeons 2. Evaluations can be run in a short amount of time, making it a feasible method in an iterative design process. Future research will investigate how procedural personas can be used as interactive inspirational tools in the content creation process and as automated critics in procedural content generation. Future work should also focus on ways to scale the procedural persona framework to games of larger complexity and on ways in which personas can learn from demonstration instead of having their utility functions specified directly.

\section*{Acknowledgment}
We thank Abdallah Saffidine and Ahmed Khalifa for advice on tree search and evolving MCTS. Michael Green acknowledges financial support from the GAANN Program.

\bibliographystyle{IEEEtran}
\bibliography{bigliography}

\end{document}